\documentclass[10pt,twocolumn,letterpaper]{article}

\usepackage[pagenumbers]{cvpr} 

\usepackage{graphicx}
\usepackage{amsmath}
\usepackage{amssymb}
\usepackage{booktabs}

\usepackage{tabu}
\usepackage{float}
\usepackage{makecell}
\usepackage{threeparttable}
\usepackage{bm}
\usepackage{xcolor}
\usepackage[lined,boxed,vlined,commentsnumbered]{algorithm2e}
\usepackage[accsupp]{axessibility}  

\usepackage[pagebackref,breaklinks,colorlinks]{hyperref}

\usepackage[capitalize]{cleveref}
\crefname{section}{Sec.}{Secs.}
\Crefname{section}{Section}{Sections}
\Crefname{table}{Table}{Tables}
\crefname{table}{Tab.}{Tabs.}

\def\cvprPaperID{7305} 
\def\confName{CVPR}
\def\confYear{2023}

\begin{document}

\title{DynamicDet: A Unified Dynamic Architecture for Object Detection}

\author{Zhihao Lin \quad Yongtao Wang\footnotemark[2] \quad  Jinhe Zhang \quad Xiaojie Chu\\
Wangxuan Institute of Computer Technology, Peking University\\
{\tt\small linzhihao@stu.pku.edu.cn, wyt@pku.edu.cn}\\
{\tt\small jinhezhang17@gmail.com, chuxiaojie@stu.pku.edu.cn}
}

\maketitle

\renewcommand{\thefootnote}{\fnsymbol{footnote}}
\footnotetext[2]{Corresponding author.}

\vspace{-1mm}
\begin{abstract}
\vspace{-1mm}
Dynamic neural network is an emerging research topic in deep learning. With adaptive inference, dynamic models can achieve remarkable accuracy and computational efficiency. However, it is challenging to design a powerful dynamic detector, because of no suitable dynamic architecture and exiting criterion for object detection. To tackle these difficulties, we propose a dynamic framework for object detection, named DynamicDet. Firstly, we carefully design a dynamic architecture based on the nature of the object detection task. Then, we propose an adaptive router to analyze the multi-scale information and to decide the inference route automatically. We also present a novel optimization strategy with an exiting criterion based on the detection losses for our dynamic detectors. Last, we present a variable-speed inference strategy, which helps to realize a wide range of accuracy-speed trade-offs with only one dynamic detector. Extensive experiments conducted on the COCO benchmark demonstrate that the proposed DynamicDet achieves new state-of-the-art accuracy-speed trade-offs. 
For instance, with comparable accuracy, the inference speed of our dynamic detector Dy-YOLOv7-W6 surpasses YOLOv7-E6 by 12\%, YOLOv7-D6 by 17\%, and YOLOv7-E6E by 39\%.
The code is available at \url{https://github.com/VDIGPKU/DynamicDet}.
\end{abstract}

\begin{figure}[t]
  \centering
   \includegraphics[width=0.92\linewidth]{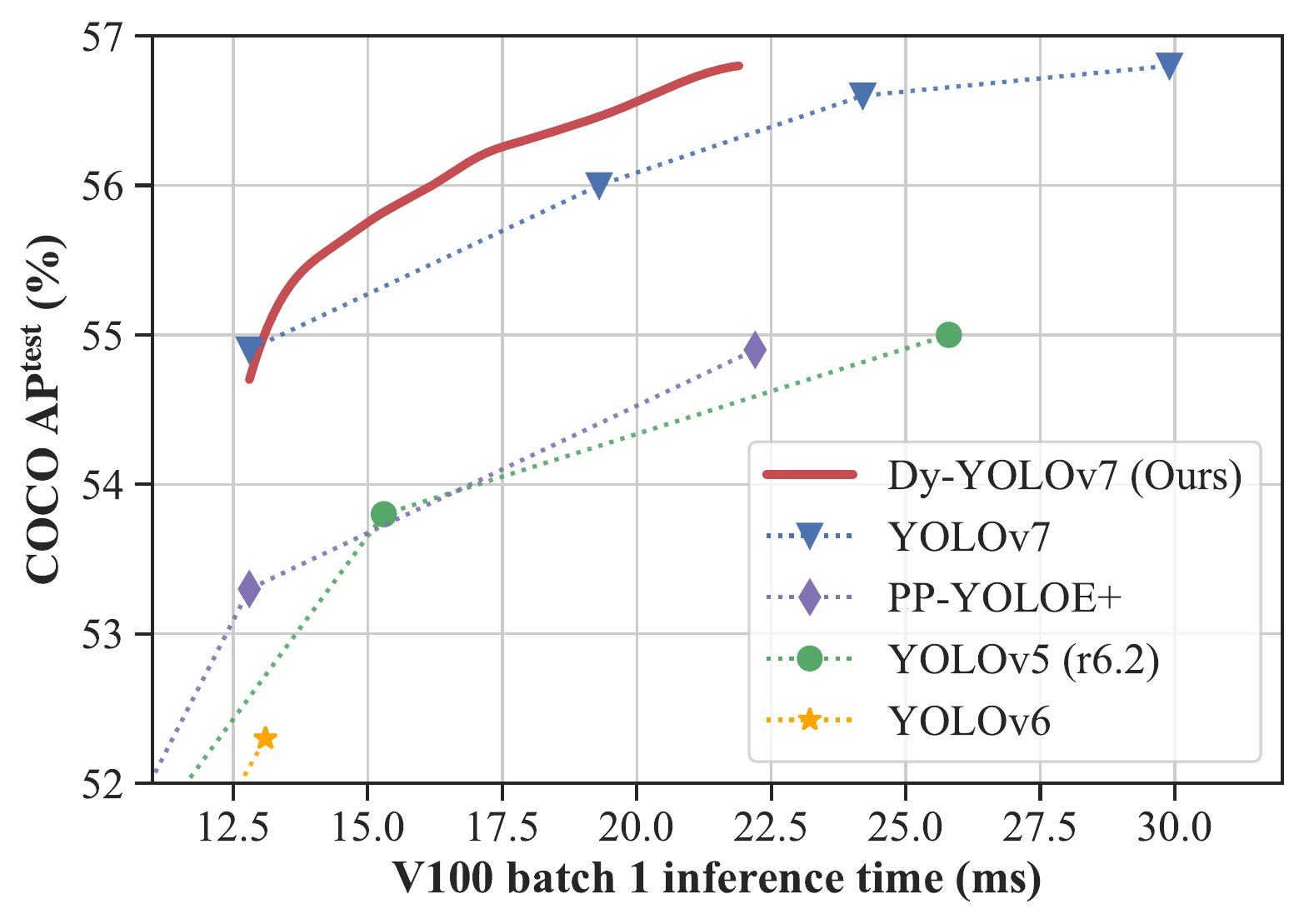}
   \vspace{-5pt}
   \caption{Comparison of the proposed dynamic detectors and other efficient object detectors. Our method can achieve a wide range of state-of-the-art trade-offs between accuracy and speed with a single model.}
   \label{fig:trade_off1}
    \vspace{-5pt}
\end{figure}

\begin{figure}[t]
  \centering
   \includegraphics[width=0.88\linewidth]{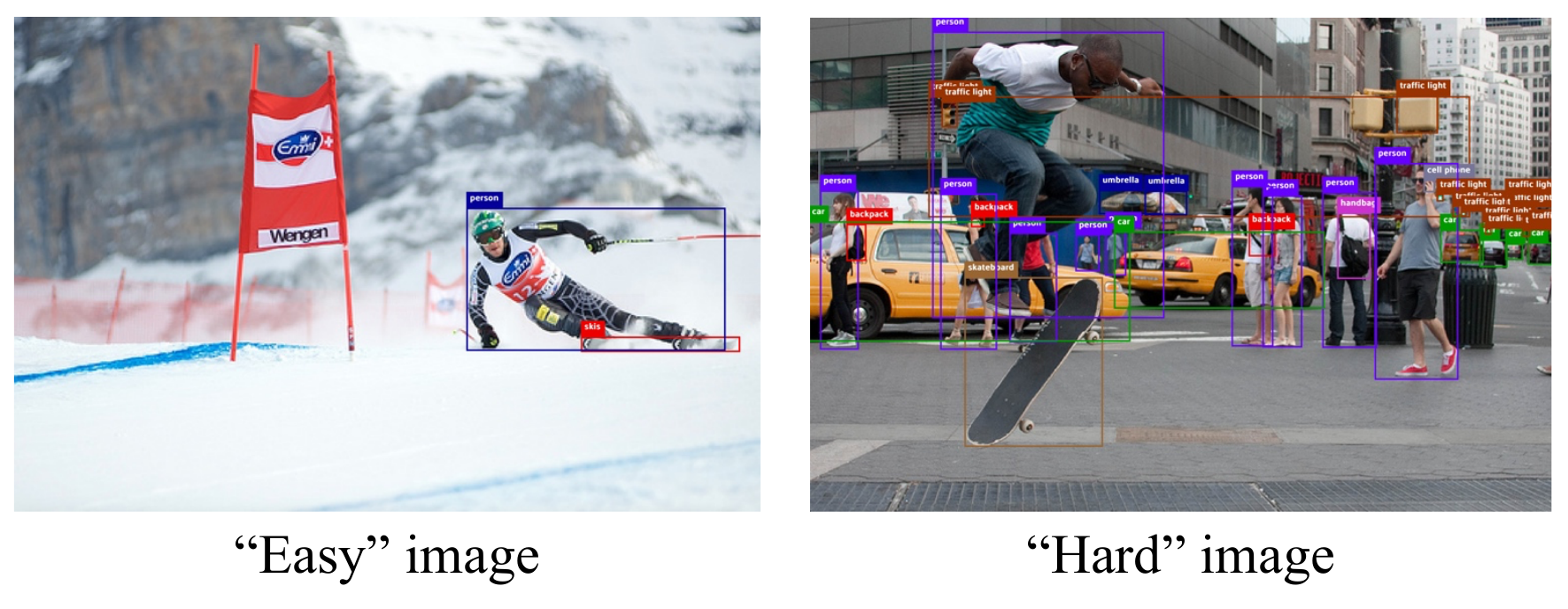}
   \vspace{-7pt}
   \caption{Examples of ``easy" and ``hard" images for the object detection task.}
   \vspace{-6mm}
   \label{fig:vis_intro}
\end{figure}

\vspace{-5.5mm}
\section{Introduction}
\vspace{-1.5mm}
Object detection is an essential topic in computer vision, as it is a fundamental component for other vision tasks, \eg, autonomous driving~\cite{shi2020pv,yang20203dssd,liang2022bevfusion}, multi-object tracking~\cite{xu2019spatial,zhang2022bytetrack}, intelligent transportation~\cite{qiu2021deep,yang2022multifeature}, \etc. In recent years, tremendous progress has been made toward more accurate and faster detectors, such as Network Architecture Search (NAS)-based detectors~\cite{ghiasi2019fpn,liang2021opanas,wang2022eautodet} and YOLO series models~\cite{bochkovskiy2020yolov4,wang2021scaled,2022githubyolov5,ge2021yolox,li2022yolov6,wang2022yolov7}.
However, these methods need to design and train multiple models to achieve a few good trade-offs between accuracy and speed, which is not flexible enough for various application scenarios.
To alleviate this problem,  we focus on dynamic inference for the object detection task, and attempt to use only one dynamic detector to achieve a wide range of good accuracy-speed trade-offs, as shown in \cref{fig:trade_off1}.

The human brain inspires many fields of deep learning, and the dynamic neural network~\cite{han2021dynamic} is a typical one.
As two examples shown in \cref{fig:vis_intro}, we can quickly identify all objects on the left ``easy" image, while we need more time to achieve the same effect for the right one.
In other words, the processing speeds of images are different in our brains~\cite{hubel1962receptive,murata2000selectivity}, which depend on the difficulties of the images. This property motivates the \textit{image-wise} dynamic neural network, and many exciting works have been proposed (\eg, Branchynet~\cite{teerapittayanon2016branchynet}, MSDNet~\cite{huang2018multiscale}, DVT~\cite{wang2021not}).
Although these approaches have achieved remarkable performance, they are all designed specifically for the image classification task and are not suitable for other vision tasks, especially for the object detection~\cite{han2021dynamic}.
The main difficulties in designing an \textit{image-wise} dynamic detector are as follows.

\textbf{Dynamic detectors cannot utilize the existing dynamic architectures.} Most existing dynamic architectures are cascaded with multiple stages (\ie, a stack of multiple layers)~\cite{mcgill2017deciding,huang2018multiscale,jie2019anytime,yang2020resolution}, and predict whether to stop the inference at each exiting point. Such a paradigm is feasible in image classification but is ineffective in object detection, since an image has multiple objects and each object usually has different categories and scales, as shown in \cref{fig:vis_intro}.
Hence, almost all detectors depend heavily on multi-scale information, utilizing the features on different scales to detect objects of different sizes (which are obtained by fusing the multi-scale features of the backbone with a detection neck, \ie, FPN~\cite{lin2017feature}). In this case, the exiting points for detectors can only be placed behind the last stage. Consequently, the entire backbone module has to be run completely~\cite{zhou2017adaptive}, and it is impossible to achieve dynamic inference on multiple cascaded stages.

\textbf{Dynamic detectors cannot exploit the existing exiting criteria for image classification.} For the image classification task, the threshold of top-1 accuracy is a widely used criterion for decision-making~\cite{huang2018multiscale,wang2021not}. Notably, it only needs one fully connected layer to predict the top-1 accuracy at intermediate layer, which is easy and costless.
However, object detection task requires the neck and the head to predict the categories and locations of the object instances~\cite{lin2017feature,ren2015faster,he2017mask,cai2018cascade}. Hence, the existing exiting criteria for image classification is not suitable for object detection.

To deal with the above difficulties, we propose a dynamic framework to achieve dynamic inference for object detection, named DynamicDet. Firstly, We design a dynamic architecture for the object detection task, which can exit with multi-scale information during the inference. Then, we propose an adaptive router to choose the best route for each image automatically. Besides, we present the corresponding optimization and inference strategies for the proposed DynamicDet.

Our main contributions are as follows:
\begin{itemize}
    \vspace{-5pt}\item We propose a dynamic architecture for object detection, named DynamicDet, which consists of two cascaded detectors and a router. This dynamic architecture can be easily adapted to mainstream detectors, \eg, Faster R-CNN and YOLO.
    \vspace{-5pt}\item We propose an adaptive router to predict the difficulty scores of the images based on the multi-scale features, and achieve automatic decision-making. In addition, we propose a hyperparameter-free optimization strategy and a variable-speed inference strategy for our dynamic architecture.
    \vspace{-5pt}\item Extensive experiments show that DynamicDet can obtain a wide range of accuracy-speed trade-offs with only one dynamic detector. We also achieve new state-of-the-art trade-offs for real-time object detection (\ie, 56.8\% AP at 46 FPS).
\end{itemize}

\vspace{-3mm}
\section{Related work}
\vspace{-0.8mm}
\subsection{Backbone design on object detection}
\vspace{-1mm}
Backbones play a crucial role in object detectors since the performance of detectors highly relies on the multi-scale features extracted by the backbones~\cite{chen2019detnas}. ResNet~\cite{he2016deep} and its variants~(\eg, ResNeXt~\cite{xie2017aggregated}, Res2Net~\cite{gao2019res2net}) introduce the residual connection to neural networks, providing a high-quality backbone architecture family for all vision tasks.
Further, to reduce the calculation load, CSPNet~\cite{wang2020cspnet} cuts down the duplicate gradient information to reduce the heavy inference, improving the efficiency significantly. Its effective architecture also inspires many lightweight detectors~(\eg, YOLO series models~\cite{bochkovskiy2020yolov4,wang2021scaled,2022githubyolov5,ge2021yolox,li2022yolov6,wang2022yolov7}).
Then, some transformer-based backbones (\eg, PVT~\cite{wang2021pyramid}, Swin Transformer~\cite{liu2021swin}) are proposed to learn the global information better. In addition, many auto-designed backbones~\cite{chen2019detnas,jiang2020sp,du2020spinenet,sun2022mae} for object detection are proposed. For example, DetNAS~\cite{chen2019detnas} utilizes the one-shot supernet to search the optimal backbone, with the guidance of the object detection task.

Although many kinds of backbones have been proposed, almost all of them are single-pass architectures, which sequentially produce one set of multi-scale features. Thus, all stages of them cannot be skipped. Fortunately, some works propose the architectures of multiple cascaded backbones, which have the potential to be converted as a dynamic backbone for object detection. For example, CBNet~\cite{liu2020cbnet,liang2022cbnet} groups multiple identical backbones with composite connections, constructing a more powerful composite backbone. Since these backbones have multiple sub-backbones and each of them can produce intermediate multi-scale features, we can add some exiting points after each sub-backbone for dynamic
inference.

\begin{figure*}[t]
  \centering
   \includegraphics[width=0.84\linewidth]{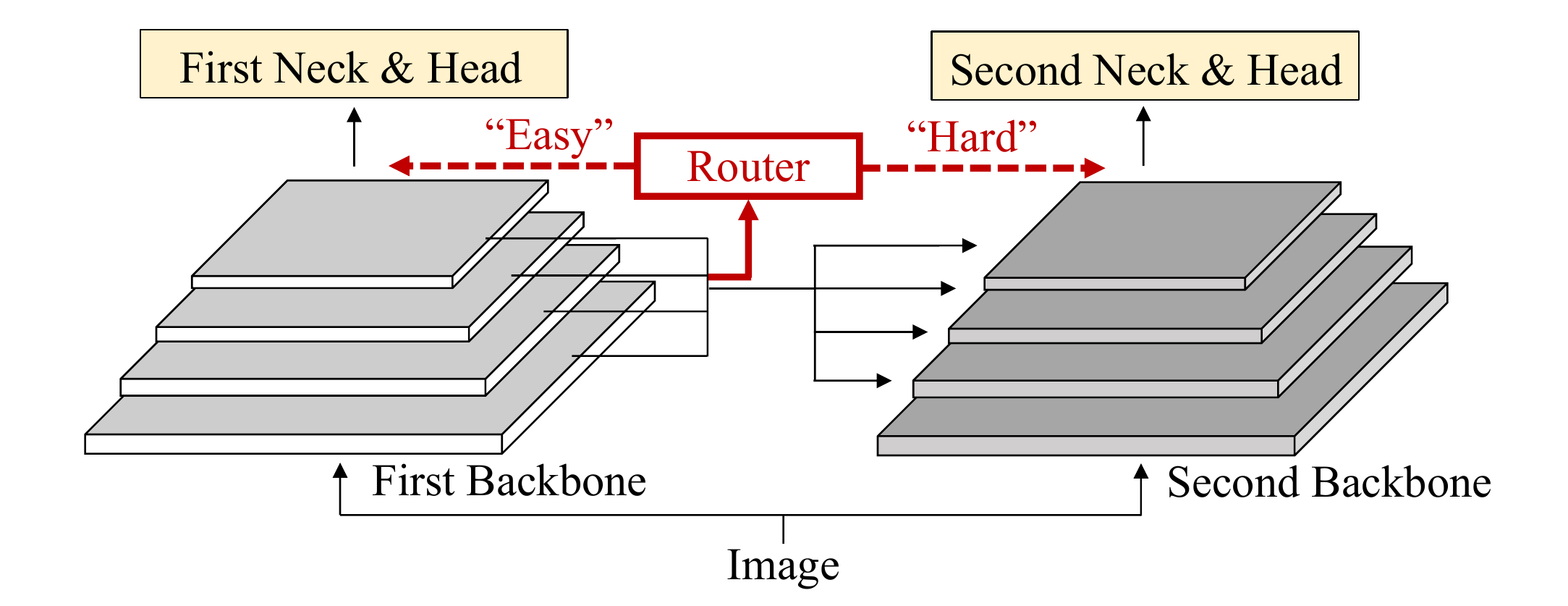}
   \vspace{-3.2mm}
   \caption{Illustration of the architecture of DynamicDet. The first backbone extracts the multi-scale features from the input image firstly. Then, the router will be fed with these multi-scale features to predict the difficulty of image and to decide the inference route. Notably, the ``easy" images will be processed by only one backbone, while the ``hard" images will be processed by two.}
   \vspace{-5.4mm}
   \label{fig:arch}
\end{figure*}

\vspace{-0.8mm}
\subsection{Accuracy-speed trade-off on object detection}
\vspace{-1mm}
Almost all detection methods are designed for a better accuracy-speed trade-off, \ie, more accurate and faster. With a given detector, the simplest way to obtain an accuracy-speed trade-off is to adopt the model scaling techniques~\cite{tan2020efficientdet,wang2021scaled,wang2022yolov7} (\eg, increasing the channel size or repeating the layers). EfficientDet~\cite{tan2020efficientdet} uniformly scales the resolution, depth, and width for all modules simultaneously, achieving remarkable efficiency on real-time detectors. Scaled-YOLOv4~\cite{wang2021scaled} modifies not only the depth, width, and resolution but also the structure of the network to pursue a better trade-off. YOLOv7~\cite{wang2022yolov7} designs a compound scaling method for concatenation-based models, achieving new state-of-the-art trade-offs.
EAutoDet~\cite{wang2022eautodet} constructs a supernet and adopts Network Architecture Search~(NAS) to automatically search for suitable scaling factors under different hardware constraints.
However, all the above methods need to train multiple detectors for the best trade-offs (\eg, one tiny model for real-time detection and another large model for accurate detection), leading to colossal training resources. In this paper, we focus on dynamic inference, aiming to achieve a wide range of best accuracy-speed trade-offs with only one dynamic detector.

\vspace{-1mm}
\subsection{Dynamic neural network}
\vspace{-1mm}
The dynamic neural network can achieve adaptive computation for different images (\ie, \textit{image-wise}~\cite{teerapittayanon2016branchynet,wang2018skipnet,huang2018multiscale,li2019improved,jie2019anytime,yang2020resolution,wang2021not,li2021dynamic}) or pixels (\ie, \textit{spatial-wise}~\cite{figurnov2017spatially,rao2021dynamicvit,han2022latency}). SACT~\cite{figurnov2017spatially} is a classic \textit{spatial-wise} dynamic network, which adaptively adjusts the number of executed layers for the regions of the image, to improve the efficiency of networks. Its practical speed-up performance highly relies on the hardware-software co-design~\cite{han2021dynamic}. However, the current deep learning hardware and libraries~\cite{abadi2016tensorflow,paszke2019pytorch} are not friendly to these \textit{spatial-wise} dynamic networks~\cite{han2021dynamic}. On the contrary, \textit{image-wise} dynamic networks do not rely on sparse computing and can be easily accelerated on the conventional CPUs and GPUs~\cite{choquette2021nvidia}. Branchynet~\cite{teerapittayanon2016branchynet} introduces the early exiting strategy, which enables the model to exit from the intermediate layer whenever the model is confident enough. MSDNet~\cite{huang2018multiscale} and its variants~\cite{li2019improved,yang2020resolution} develop a multi-classifier architecture for the image classification task. DVT~\cite{wang2021not} cascades multiple transformers with increasing numbers of tokens and activates them sequentially to achieve dynamic inference. However, these methods are all designed specifically for the image classification task and cannot be applied to other vision tasks, such as object detection.

The closest work to our DynamicDet is Adaptive Feeding~\cite{zhou2017adaptive}. In Adaptive Feeding~\cite{zhou2017adaptive}, each image is detected by a lightweight detector~(\eg, Tiny YOLO~\cite{redmon2016you}) and then classified as easy or hard by a linear support vector machine~(SVM) with those detected results. Then, the easy images will go through a fast detector~(\eg, SSD300~\cite{liu2016ssd}), while the hard images will go through a more accurate but slower one~(\eg, SSD500~\cite{liu2016ssd}). Adaptive Feeding~\cite{zhou2017adaptive} introduces the above multi-stage process for dynamic inference, which is inefficient and not elegant. In comparison, the proposed DynamicDet cascades two detectors and a classifier~(\ie, the router), yielding a more unified and efficient dynamic detector.

\vspace{-0.6mm}
\section{Approach}
\vspace{-0.6mm}
In the following, we elaborate on our dynamic architecture for object detection. We first introduce the overall architecture in \cref{sec:overall_architecture}. Then, we state the proposed adaptive router, \ie, the decision maker of DynamicDet in \cref{sec:amsr}. Finally, we introduce the optimization strategy and a variable-speed inference strategy in \cref{sec:train,sec:inference}.

\vspace{-0.6mm}
\subsection{Overall architecture}
\label{sec:overall_architecture}
\vspace{-0.6mm}
The overall architecture of our dynamic detector is shown in \cref{fig:arch}. 
Inspired by CBNet~\cite{liu2020cbnet,liang2022cbnet}, our dynamic architecture consists of two detectors and one router. For an input image $\mathbf{x}$, we initially extract its multi-scale features $F_1$ with the first backbone $\mathcal{B}_1$ as
\vspace{-0.6mm}
\begin{equation}
\small
F_1 =\mathcal{B}_1(\mathbf{x})=[f_1^{\{1\}}, f_1^{\{2\}}, \dots, f_1^{\{L\}}],\vspace{-0.6mm}
\end{equation}
where $L$ denotes the number of stages, \ie, the number of multi-scale features.
Then, the router $\mathcal{R}$ will be fed with these features $F_1$ to predict a difficulty score $\phi\in(0, 1)$ for this image as
\vspace{-0.6mm}
\begin{equation}
\small
\phi = \mathcal{R}(F_1).\vspace{-0.6mm}
\end{equation}
Generally speaking, the ``easy" images exit at the first backbone, while the ``hard" images require the further processing.
Specifically, if the router classifies the input image as an ``easy" one,
the followed neck and head $\mathcal{D}_1$ will output the detection results $\mathbf{y}$ as
\vspace{-0.6mm}
\begin{equation}
\small
\mathbf{y} = \mathcal{D}_1(F_1).\vspace{-0.6mm}
\end{equation}
On the contrary, if the router classifies the input image as a ``hard" one, the multi-scale features will need further enhancement by the second backbone, instead of immediately decoded by $\mathcal{D}_1$.
In particular, we embed the multi-scale features $F_1$ into $H$ by a composite connection module $\mathcal{G}$ as
\vspace{-0.6mm}
\begin{equation}
\small
\label{eq:reuse_module}
H = \mathcal{G}(F_1)= [h^{\{1\}}, h^{\{2\}}, \dots, h^{\{L\}}],\vspace{-0.6mm}
\end{equation}
where $\mathcal{G}$ is the DHLC of CBNet~\cite{liu2020cbnet,liang2022cbnet} in our implementation.
Then, we feed the input image $\mathbf{x}$ into the second backbone and enhance the features of the second backbone via summing the corresponding elements of $H$ at each stage sequentially, denoted as
\vspace{-0.6mm}
\begin{equation}
\small
F_2 = \mathcal{B}_2(\mathbf{x}, H) = [f_2^{\{1\}}, f_2^{\{2\}}, \dots, f_2^{\{L\}}],\vspace{-0.6mm}
\end{equation}
and the detection results will be obtained by the second head and neck $\mathcal{D}_2$ as
\vspace{-0.6mm}
\begin{equation}
\small
\mathbf{y} = \mathcal{D}_2(F_2).\vspace{-0.6mm}
\end{equation}

Through the above process, the ``easy'' images will be processed by only one backbone, while the ``hard" images will be processed by two.
Obviously, with such an architecture, trades-offs between computation (\ie, speed) and accuracy can be achieved.

\vspace{-0.6mm}
\subsection{Adaptive router}
\label{sec:amsr}
\vspace{-0.6mm}
In mainstream object detectors, different scale features play different roles.
Generally, the features of the shallow layers, with strong spatial information and small receptive fields, are more used to detect small objects. In contrast, the features of the deep layers, with strong semantic information and large receptive fields, are more used to detect large objects.
This property makes it necessary to consider multi-scale information when predicting the difficulty score of an image.
According to this, we design an adaptive router based on the multi-scale features, that is, a simple yet effective decision-maker for the dynamic detector. 

Inspired by the squeeze-and-excitation~(SE) module~\cite{hu2018squeeze}, we first pool the multi-scale features $F_1$ independently and concatenate them all as
\vspace{-0.6mm}
\begin{equation}
\small
\widetilde{F}_1=\mathcal{C}([\mathcal{P}(f_1^{\{1\}}), \mathcal{P}(f_1^{\{2\}}), \dots, \mathcal{P}(f_1^{\{L\}})]),\vspace{-0.6mm}
\end{equation}
where $\mathcal{P}$ denotes the global average pooling and $\mathcal{C}$ denotes the channel-wise concatenation.
With this operation, we compress the multi-scale features $F_1$ into a vector  $\widetilde{F}_1\in\mathbb{R}^{d}$ of dimension $d$. Then, we map this vector to a difficulty score $\phi\in(0, 1)$ via two learnable fully connected layers as 
\vspace{-2.5mm}
\begin{equation}
\label{eq:router}
\small
\phi = \sigma(W_2(\delta(W_1\widetilde{F}_1 + b_1))+b_2),\vspace{-0.6mm}
\end{equation}
where $\delta, \sigma$ denote the ReLU and Sigmoid activation functions respectively, and $W_1, W_2, b_1, b_2$ are learnable parameters. Following \cite{zhu2020dynamic}, we reduce the feature dimension to $\lfloor d/4\rfloor$ in the first fully connected layer, and exploit the second fully connected layer with a Sigmoid function to generate the predicted score.
It is worth noting that the computational burden of our router can be negligible since we first pool all multi-scale features to one vector.

\begin{figure*}[t]
  \centering
   \includegraphics[height=0.35\linewidth]{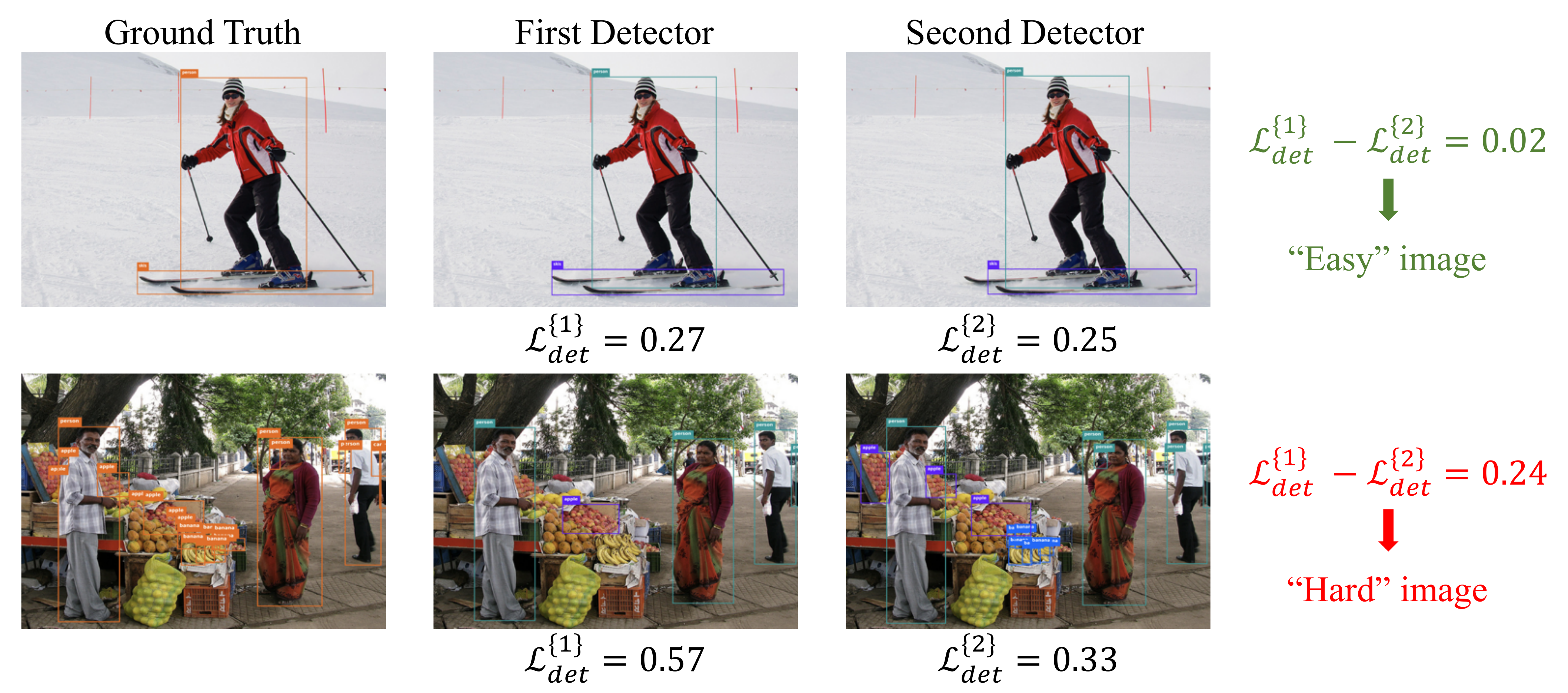}
   \vspace{-1.5mm}
   \caption{Illustration of the difficulty criterion based on the training loss difference between two cascaded detectors. For the top image, the loss difference between the first detector and the second detector is very small, so it should be classified as an ``easy" image. On the contrary, the loss difference of the bottom image is large, so it should be classified as a ``hard" image.}
   \vspace{-2mm}
   \label{fig:lossfig}
\end{figure*}

\subsection{Optimization strategy}
\label{sec:train}
\begin{figure}[t]
  \centering
   \includegraphics[height=0.5\linewidth]{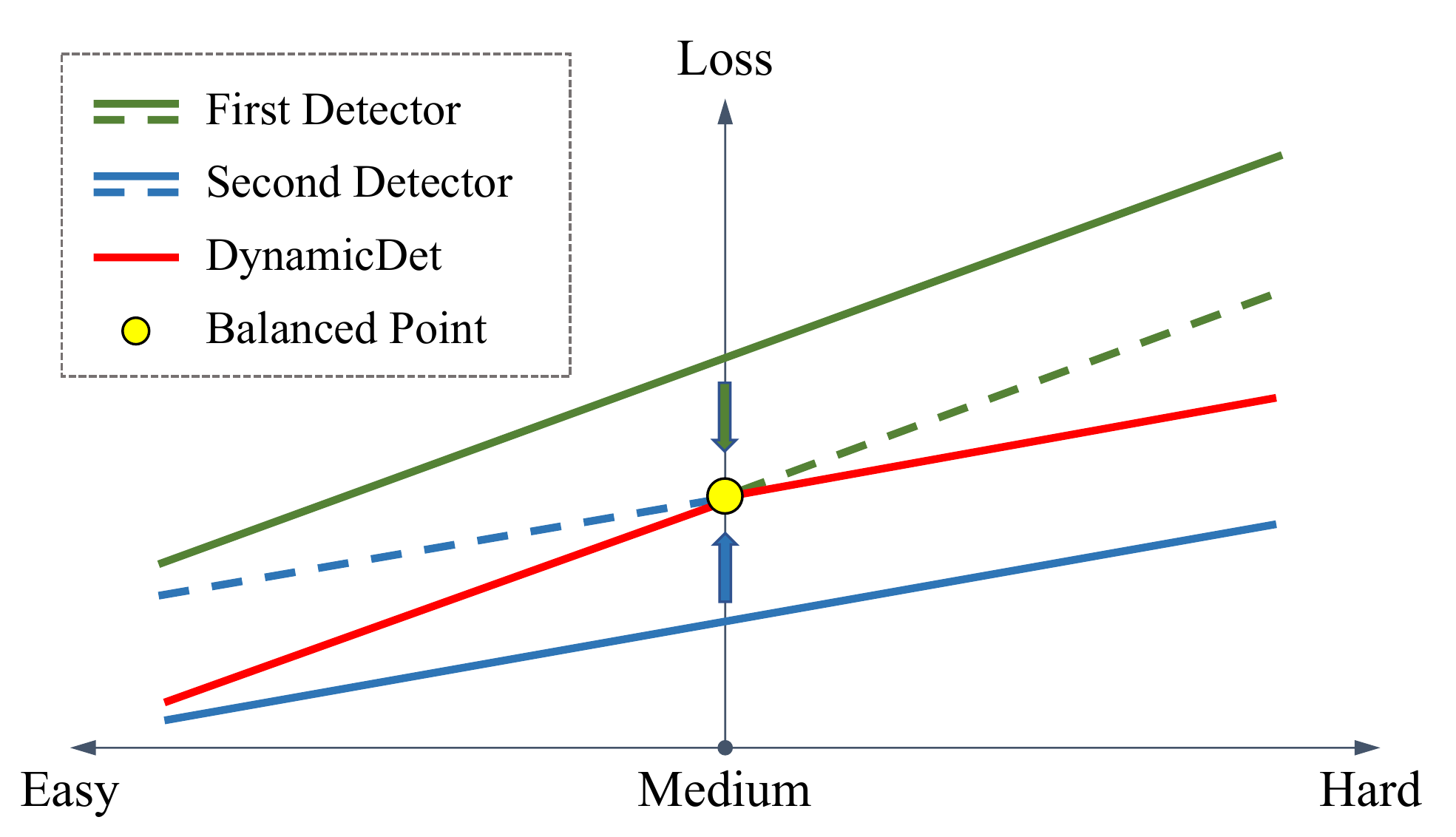}
   \vspace{-1.5mm}
   \caption{Qualitative analysis of the loss of two cascaded detectors on the images with different difficulties. With the proposed adaptive offset in our optimization strategy, the loss curves of two detectors intersect and reveal the optimal curve of DynamicDet.}
   \label{fig:loss}
   \vspace{-2mm}
\end{figure}
In this section, we describe the optimization strategy for the above dynamic architecture. 

Firstly, we jointly train the cascaded detectors, and the training objective is
\vspace{-0.6mm}
\begin{equation}
\small
\min_{\Theta_1, \Theta_2}\ \ (\mathcal{L}_{det}^{\{1\}}(\mathbf{x}, \mathbf{y}|\Theta_1) + \mathcal{L}_{det}^{\{2\}}(\mathbf{x}, \mathbf{y}|\Theta_2)),\vspace{-0.6mm}
\end{equation}
where $\mathbf{x}, \mathbf{y}$ denote the input image and the ground truth respectively, $\Theta_i$ denotes the learnable parameters of the detector $i$ and $\mathcal{L}_{det}^{\{i\}}$ denotes the training loss for detector $i$ (\eg, bounding box regression loss and classification loss). After the above training phase, these two detectors will be able to detect the objects, and we freeze their parameters $\Theta_1, \Theta_2$ during the later training.

Then, we train the adaptive router to automatically distinguish the difficulty of the image.
Here, we assume the parameters of the router are $\Theta_{\mathcal{R}}$ and the predicted difficulty score obtained from the \cref{eq:router} is $\phi$.
We hope the router can assign the ``easy" images~(\ie, with lower $\phi$) to the faster detector (\ie, the first detector) and the ``hard" images~(\ie, with higher $\phi$) to the more accurate detector (\ie, the second detector).

However, it is non-trivial to implement that in practice. If we directly optimize the router without any constraints as
\vspace{-4mm}
\begin{align}
\small
\begin{split}
\min_{\Theta_{\mathcal{R}}}\ \ ( (1-\phi)\mathcal{L}_{det}^{\{1\}}(\mathbf{x}, \mathbf{y}|\Theta_1) + \phi\mathcal{L}_{det}^{\{2\}}(\mathbf{x}, \mathbf{y}|\Theta_2)),\vspace{-1.5mm}
\end{split}
\end{align}
the router will always choose the most accurate detector as it allows for a lower training loss.
Furthermore, if we naively add hardware constraints to the training objective as
\vspace{-1mm}
\begin{align}
\small
\begin{split}
\min_{\Theta_{\mathcal{R}}}\ \ ( (1-&\phi)\mathcal{L}_{det}^{\{1\}}(\mathbf{x}, \mathbf{y}|\Theta_1) \\ &+ \phi\mathcal{L}_{det}^{\{2\}}(\mathbf{x}, \mathbf{y}|\Theta_2) \textcolor{red}{+\bm{\lambda\phi}}),\vspace{-4mm}
\end{split}
\end{align}
we will have to adjust the hyperparameter $\lambda$ by try and error, leading to huge workforce consumption.

To overcome the above challenges, we propose a hyperparameter-free optimization strategy for our adaptive router. 
First, we define the difficulty criterion based on the corresponding training loss difference between two detectors of an image, as shown in \cref{fig:lossfig}. Specifically, we assume that if the loss difference of an image between two detectors is small enough, this image can be classified as an ``easy" image. Instead, if the loss difference is large enough, it should be classified as a ``hard" image.
Ideally, for a balanced situation, we hope the easier half of all images go through the first detector, and the harder half go through the second one.
To achieve this, we introduce an adaptive offset to balance the losses of two detectors and optimize our router via gradient descent.
In practice, we first calculate the median of the training loss difference $\Delta$ between the first and the second detector on the training set. 
Then, the training objective of our router can be formulated as
\begin{align}
\small
\begin{split}
\min_{\Theta_{\mathcal{R}}}\ \ ((1-&\phi)(\mathcal{L}_{det}^{\{1\}}(\mathbf{x}, \mathbf{y}|\Theta_1) \textcolor{red}{-\bm{\Delta /2}})\\ & + \phi(\mathcal{L}_{det}^{\{2\}}(\mathbf{x}, \mathbf{y}|\Theta_2) \textcolor{red}{+\bm{\Delta /2}})),
\end{split}
\end{align}
where $\Delta/2$ is used to reward the first detector and punish the second detector, respectively.
As the qualitative analysis shown in \cref{fig:loss}, without this reward and penalty, the losses of the second detector are always smaller than the first detector.
When the reward and penalty are conducted, their loss curves intersect and reveal the optimal curve.

Our training objective provides a means to optimize the adaptive router by introducing the following gradient through the difficulty score $\phi$ to all parameters $\Theta_{\mathcal{R}}$ of the router as
\vspace{-0.6mm}
\begin{equation}
\label{eq:gradient}
\small
\frac{\partial\mathcal{L}}{\partial\Theta_{\mathcal{R}}} = \frac{\partial\mathcal{L}}{\partial\phi} \frac{\partial\phi}{\partial \Theta_{\mathcal{R}}} =-\frac{\partial\phi}{\partial \Theta_{\mathcal{R}}}\textcolor{red}{\bm{(\mathcal{L}_{det}^{\{1\}} - \mathcal{L}_{det}^{\{2\}}- \Delta)}}.\vspace{-0.6mm}
\end{equation}
To distinguish between ``easy" and ``hard" images better, we expect the optimization direction of the router to be related to the difficulty of the image, \ie, the difference in loss between the two detectors. Obviously, the gradient at \cref{eq:gradient} enable such expectation.

\begin{table}[ht]
\centering
\addtolength{\tabcolsep}{0.8pt}
\small
\begin{threeparttable}
\begin{tabular}{lrrrr}
\toprule
Model                       & Size          & FLOPs  & FPS            & AP            \\
\midrule
EAutoDet-X~\cite{wang2022eautodet}                  & 640           & 225.3G & 41\tnote{\dag}             & 49.2          \\
\midrule
YOLOX-L~\cite{ge2021yolox}                     & 640           & 155.6G & 69\tnote{\dag}             & 50.1          \\
YOLOX-X~\cite{ge2021yolox}                     & 640           & 281.9G & 58\tnote{\dag}             & 51.5          \\
\midrule
YOLOv5-L (r6.2)~\cite{2022githubyolov5}             & 640           & 109.1G & 114            & 49.0          \\
YOLOv5-X (r6.2)~\cite{2022githubyolov5}             & 640           & 205.7G & 100            & 50.9          \\
\midrule
YOLOv6-M~\cite{li2022yolov6}                    & 640           & 82.2G  & 109            & 49.6          \\
YOLOv6-L~\cite{li2022yolov6}                    & 640           & 144.0G & 76             & 52.4          \\
\midrule
PP-YOLOE+-M~\cite{xu2022pp}                 & 640           & 49.9G  & 123\tnote{\dag}            & 50.0          \\
PP-YOLOE+-L~\cite{xu2022pp}                 & 640           & 110.1G & 78\tnote{\dag}             & 53.3          \\
PP-YOLOE+-X~\cite{xu2022pp}                 & 640           & 206.6G & 45\tnote{\dag}             & 54.9          \\
\midrule
YOLOv7~\cite{wang2022yolov7}                      & 640           & 104.7G & 114            & 51.4          \\
\textbf{Dy-YOLOv7 / 10}     & \textbf{640}  & \textbf{112.4G} & \textbf{110} & \textbf{52.1} \\
\textbf{Dy-YOLOv7 / 50}     & \textbf{640}  & \textbf{143.2G} & \textbf{96}  & \textbf{53.3} \\
\textbf{Dy-YOLOv7 / 90}     & \textbf{640}  & \textbf{174.0G} & \textbf{85}  & \textbf{53.8} \\
\textbf{Dy-YOLOv7 / 100}    & \textbf{640}  & \textbf{181.7G} & \textbf{83}  & \textbf{53.9} \\
\midrule
YOLOv7-X~\cite{wang2022yolov7}                    & 640           & 189.9G & 105            & 53.1          \\
\textbf{Dy-YOLOv7-X / 10}   & \textbf{640}  & \textbf{201.7G} & \textbf{98}  & \textbf{53.3} \\
\textbf{Dy-YOLOv7-X / 50}   & \textbf{640}  & \textbf{248.9G} & \textbf{78}  & \textbf{54.4} \\
\textbf{Dy-YOLOv7-X / 90}   & \textbf{640}  & \textbf{296.1G} & \textbf{65}  & \textbf{55.0} \\
\textbf{Dy-YOLOv7-X / 100}  & \textbf{640}  & \textbf{307.9G} & \textbf{64}  & \textbf{55.0} \\
\midrule
\midrule
YOLOv5-M6 (r6.2)~\cite{2022githubyolov5}            & 1280          & 200.0G & 96             & 51.4          \\
YOLOv5-L6 (r6.2)~\cite{2022githubyolov5}            & 1280          & 445.6G & 65             & 53.8          \\
YOLOv5-X6 (r6.2)~\cite{2022githubyolov5}            & 1280          & 839.2G & 39             & 55.0          \\
\midrule
YOLOv7-W6~\cite{wang2022yolov7}                   & 1280          & 360.0G & 78             & 54.9          \\
YOLOv7-E6~\cite{wang2022yolov7}                   & 1280          & 515.2G & 52             & 56.0          \\
YOLOv7-D6~\cite{wang2022yolov7}                   & 1280          & 806.8G & 41             & 56.6          \\
YOLOv7-E6E~\cite{wang2022yolov7}                  & 1280          & 843.2G & 33             & 56.8          \\
\midrule
\textbf{Dy-YOLOv7-W6 / 10}  & \textbf{1280} & \textbf{384.2G} & \textbf{74}    & \textbf{55.2} \\
\textbf{Dy-YOLOv7-W6 / 50}  & \textbf{1280} & \textbf{480.8G} & \textbf{58}    & \textbf{56.1} \\
\textbf{Dy-YOLOv7-W6 / 90}  & \textbf{1280} & \textbf{577.4G} & \textbf{48}    & \textbf{56.7} \\
\textbf{Dy-YOLOv7-W6 / 100} & \textbf{1280} & \textbf{601.6G} & \textbf{46}    & \textbf{56.8} \\
\bottomrule
\end{tabular}
\begin{tablenotes}
	\footnotesize
	\item[1] The FPS marked with $\dag$ are from the corresponding papers, and others are measured on the same machine with 1 NVIDIA V100 GPU.
\end{tablenotes}
\end{threeparttable}
\caption{Comparison with the state-of-the-art real-time object detectors on COCO \textit{test-dev}.}
\vspace{-4mm}
\label{tab:results}
\end{table}

\subsection{Variable-speed inference}
\label{sec:inference}
We further propose a simple and effective method to determine the difficulty score thresholds to achieve variable-speed inference with only one dynamic detector. 
Specifically, our adaptive router will output a difficulty score and decide which detector to go through based on a certain threshold during inference.  
Therefore, we can set different thresholds to achieve different accuracy-speed trade-offs. Firstly, we count the difficulty scores $\mathcal{S}_{val}$ of the validation set. Then, based on the actual needs (\eg, the target latency), we can obtain the corresponding threshold for our router. For example, assuming the latency of the first detector is $lat_{1}$, the latency of the cascaded two detectors is $lat_{2}$ and the target latency is $lat_{t}$, we can calculate the maximum allowable proportion of the ``hard" images $k$ as
\vspace{-0.6mm}
\begin{equation}
\small
k=\frac{lat_t - lat_1}{lat_2-lat_1},\ \ \ \  lat_1\le lat_t \le lat_2,\vspace{-0.6mm}
\end{equation}
and then the threshold $\tau_{val}$ will be
\vspace{-0.6mm}
\begin{equation}
\small
\tau_{val}=\operatorname{percentile}(\mathcal{S}_{val}, k),\vspace{-0.6mm}
\end{equation}
where $\operatorname{percentile}(\cdot, k)$ means to compute the $k$-th quantile of the data. 
It is worth noting that this threshold $\tau_{val}$ is robust in both validation set and test set because these two sets are independent and identically distributed (\ie, i.i.d.).

Based on the above strategy, one dynamic detector can directly cover the accuracy-speed trade-offs from the single to double detectors, avoiding redesigning and training multiple detectors under different hardware constraints.

\vspace{-1mm}\section{Experiments}
\vspace{-1mm}
In this section, we evaluate our DynamicDet through extensive experiments. In \cref{sec:setups}, we detail the experimental setups. In \cref{sec:one-stage},  we compare our DynamicDet with the state-of-the-art real-time detectors. In \cref{sec:two-stage}, we present the experimental results on two-stage detectors with CNN- and transformer-based backbones to demonstrate the generality of DynamicDet over different backbones and detectors. In \cref{sec:ana}, we ablate each component of DynamicDet in detail. In \cref{sec:visualization}, we visualize the ``easy" and the ``hard" images determined by the adaptive router.

\vspace{-1mm}
\subsection{Experimental setups}
\label{sec:setups}
\vspace{-1mm}
We conduct experiments on the COCO~\cite{lin2014microsoft} benchmark. All the models presented are trained on the 118k training images, and tested on the 5k \textit{minival} images and 20k \textit{test-dev} images.
We choose the YOLOv7~\cite{wang2022yolov7} series models as the real-time detector baseline, and the Faster R-CNN~\cite{ren2015faster} (ResNet~\cite{he2016deep}) and the Mask R-CNN~\cite{he2017mask} (Swin Transformer~\cite{liu2021swin}) as the two-stage detector baselines. 
All dynamic detectors are trained with the same hyper-parameters of their corresponding baselines. 
We use brief notation to indicate the easy-hard proportion for each dynamic detector: for instance, ``Dy-YOLOv7-X/10" means the dynamic YOLOv7-X model with 10\% images are classified as ``hard" and the rest are classified as ``easy". 
The training of the adaptive router is conducted on a single GPU with batchsize 1 and two epochs, utilizing the AdamW~\cite{loshchilov2017decoupled} optimizer with a constant learning rate $1\times 10^{-5}$ and weight decay $5\times 10^{-3}$.
The reported FLOPs for dynamic detectors are the average FLOPs on the corresponding dataset.
The speed performance is measured on a machine with 1 NVIDIA V100 GPU unless otherwise stated.
The implementation of Dy-YOLOv7 is developed by the YOLOv7~\cite{wang2022yolov7} framework, with two identical detectors. The implementation of dynamic two-stage detectors is developed by the open-source CBNet~\cite{liang2022cbnet} framework, with two identical backbones and a shared neck and head.

\vspace{-1mm}
\subsection{Comparison with the state-of-the-arts}
\label{sec:one-stage}
\vspace{-1mm}
As shown in \cref{tab:results}, compared with the state-of-the-art high-performance real-time object detectors, our dynamic detectors obtain better results and achieve the new state-of-the-art accuracy-speed trade-offs.
Specifically, Dy-YOLOv7-W6 / 50 achieves 56.1\% AP with 58 FPS, which is 0.1\% more accurate and 12\% faster than YOLOv7-E6. Dy-YOLOv7-W6 / 100 achieves 56.8\% AP with 46 FPS, which is 39\% faster than YOLOv7-E6E with a similar accuracy. It is worth noting that \textit{these trade-offs are obtained by only one dynamic detector instead of multiple independent models}.

\vspace{-1mm}
\subsection{Generality for two-stage detectors}
\label{sec:two-stage}
\vspace{-1mm}
\begin{table}[t]
\small
\centering
\addtolength{\tabcolsep}{-4pt}
\begin{tabular}{lrrrr}
\toprule
Model  & FLOPs &FPS & AP$_\text{box}$  & AP$_\text{mask}$   \\
\midrule
Faster R-CNN ResNet50~\cite{ren2015faster,he2016deep} & 207.1G &23 & 37.4 & -    \\
Faster R-CNN ResNet101~\cite{ren2015faster,he2016deep}  & 283.1G &18  & 39.4 & -    \\
\textbf{Dy-Faster R-CNN ResNet50 / 50} & \textbf{245.4G} &\textbf{20}  & \textbf{39.5} & -    \\
\textbf{Dy-Faster R-CNN ResNet50 / 90} & \textbf{276.0G} &\textbf{17}   & \textbf{40.4} & -    \\
\midrule
Mask R-CNN Swin-T~\cite{he2017mask,liu2021swin}   & 263.8G  & 15 & 46.0 & 41.6 \\
Mask R-CNN Swin-S~\cite{he2017mask,liu2021swin}   & 353.8G   & 12   & 48.2 & 43.2 \\
\textbf{Dy-Mask R-CNN Swin-T / 50}   & \textbf{310.6G} & \textbf{12} & \textbf{48.7} & \textbf{43.6} \\
\textbf{Dy-Mask R-CNN Swin-T / 90}   & \textbf{348.0G} & \textbf{11}  & \textbf{49.9} & \textbf{44.2}\\
\bottomrule
\end{tabular}
\vspace{-1mm}
\caption{Comparison with two-stage detectors on COCO \textit{minival}.}
\vspace{-1.5mm}
\label{tab:two_stage}
\end{table}

\begin{figure}[t]
  \centering
   \includegraphics[width=0.97\linewidth,height=0.6\linewidth]{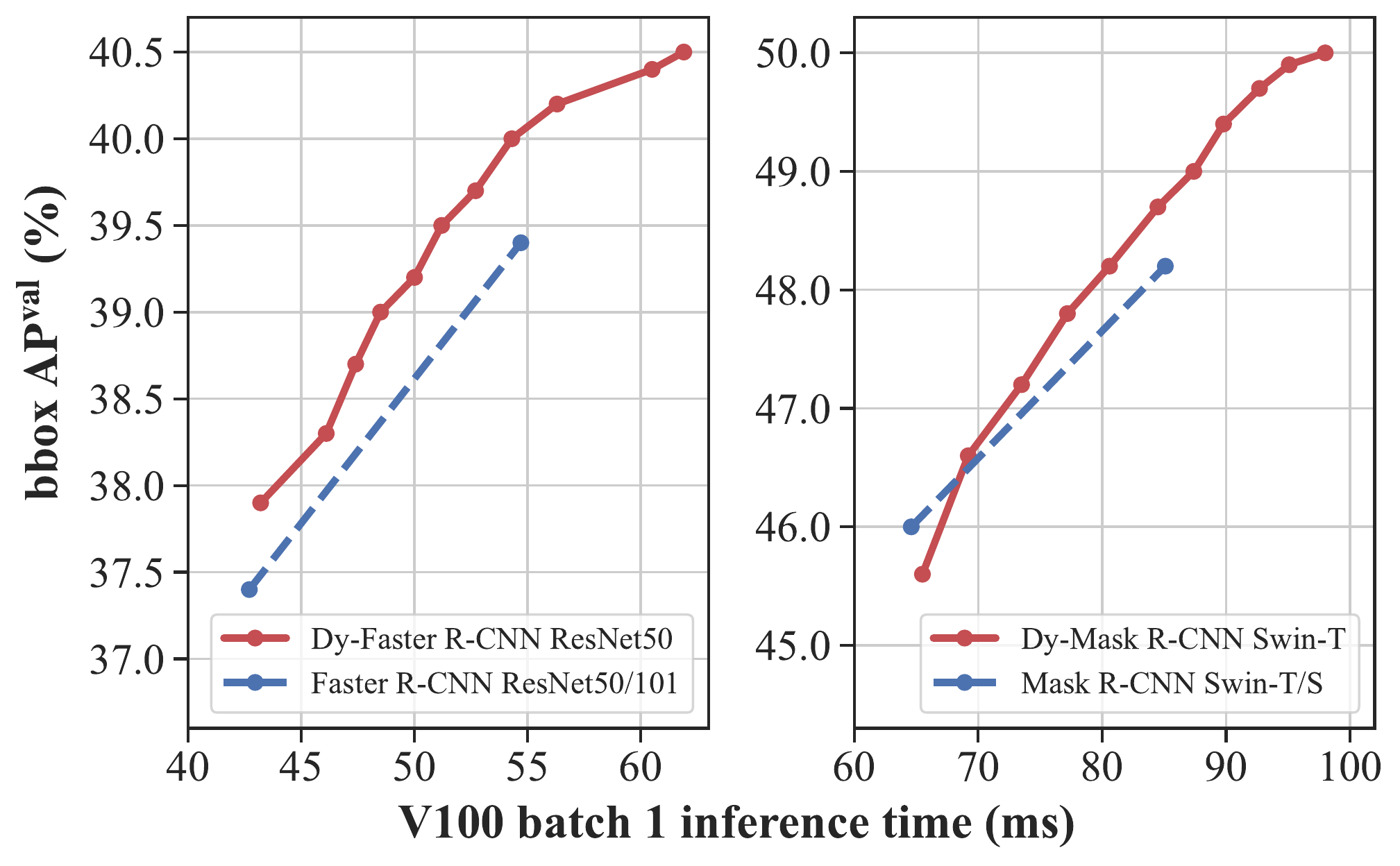}
   \vspace{-3mm}
   \caption{Bounding box mAP \textit{v.s.} inference speed for two-stage detectors on COCO \textit{minival}.}
   \vspace{-4mm}
   \label{fig:trade_off_two_stage}
\end{figure}
We conduct experiments on two classic two-stage detectors (\ie, Faster R-CNN~\cite{ren2015faster}, Mask R-CNN~\cite{he2017mask}) to show the generality of our DynamicDet. As shown in \cref{tab:two_stage}, our method is compatible with two-stage detectors and can also improve the accuracy-speed performance of baselines. For example, Dy-Faster R-CNN ResNet50 / 90 boosts the bbox AP by 1\% with the comparable inference speed for Faster R-CNN ResNet101. Furthermore, DynamicDet is also compatible with transformer-based backbones (\eg, Swin Transformer~\cite{liu2021swin}). Dy-Mask R-CNN Swin-T / 90 improves the bbox AP to 49.9\% with the comparable inference speed of Mask R-CNN Swin-S. Notably, our two-stage dynamic detector can also perform variable-speed inference as illustrated in \cref{fig:trade_off_two_stage}. 

\vspace{-1mm}
\subsection{Ablation study}
\label{sec:ana}
\subsubsection{Lightweight adaptive router}
\vspace{-1mm}
The FLOPs ratios for the adaptive router in different models are presented in \cref{tab:router_flops}. We can find that this ratio is less than 0.002\% in all models, demonstrating that the computational burden of the adaptive router can be negligible. This lightweight router avoids slowing down the detection process and ensures the fast decision-making for dynamic inference.

\vspace{-3mm}
\subsubsection{Effective training strategy for adaptive router}
\begin{figure}[t]
  \centering
   \includegraphics[height=0.6\linewidth]{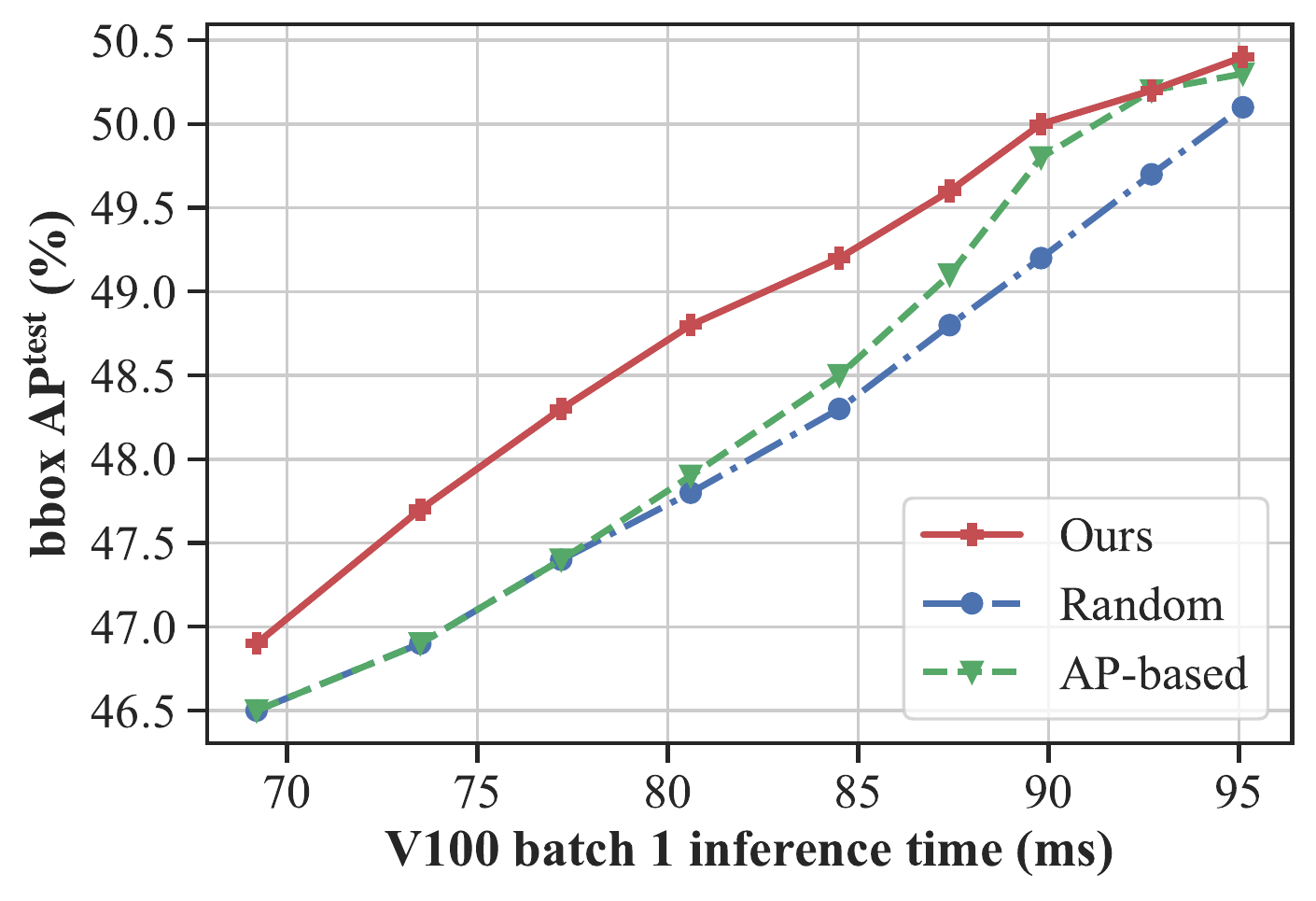}
   \vspace{-3mm}
   \caption{Comparison of the proposed strategy and other two decision-making strategies.}
   \vspace{-4mm}
   \label{fig:other_train}
\end{figure}

\vspace{-1mm}
We ablate the effectiveness of the proposed training and optimization strategy for adaptive router. We first train a Mask R-CNN~\cite{he2017mask} with cascaded Swin-T~\cite{liu2021swin} as our baseline detector. Then, we apply three strategies to achieve the decision-making for router: random, AP-based (\ie, dividing ``easy" and ``hard" images based on the validation accuracy and using them to train the router, similar to Adaptive Feeding~\cite{zhou2017adaptive}), and our proposed strategy. 
As shown in \cref{fig:other_train}, we compare the bbox AP on the \textit{test-dev} set of different training strategies. It is shown that our optimization strategy outperforms another two strategies under all latency constraints. Taking the detector with 84.5 ms latency (\ie, 50\% easy and 50\% hard) as an example, our strategy exceeds random selection 0.9\% AP and AP-based strategy 0.7\% AP.
This proves that our optimization strategy effectively improves the discrimination accuracy of the router and outperforms AP-based strategy~\cite{zhou2017adaptive}.

\begin{table}[t]
\small
\begin{tabular}{lrrr}
\toprule
Model                & Router & Total & Ratio    \\
\midrule
Dy-YOLOv7            & 2.1M   & 104.7G   & 0.0020\% \\
Dy-YOLOv7-W6         & 1.9M   & 360.0G   & 0.0005\% \\
\midrule
Dy-Faster R-CNN ResNet50  & 3.7M   & 283.7G   & 0.0013\% \\
Dy-Mask R-CNN Swin-T & 0.5M   & 357.4G   & 0.0001\% \\
\bottomrule
\end{tabular}
\vspace{-1mm}
\caption{Comparison of the adaptive router's FLOPs and the total FLOPs in different dynamic models.}
\vspace{-1mm}
\label{tab:router_flops}
\end{table}

\begin{figure*}[t]
  \centering
   \includegraphics[height=0.30\linewidth]{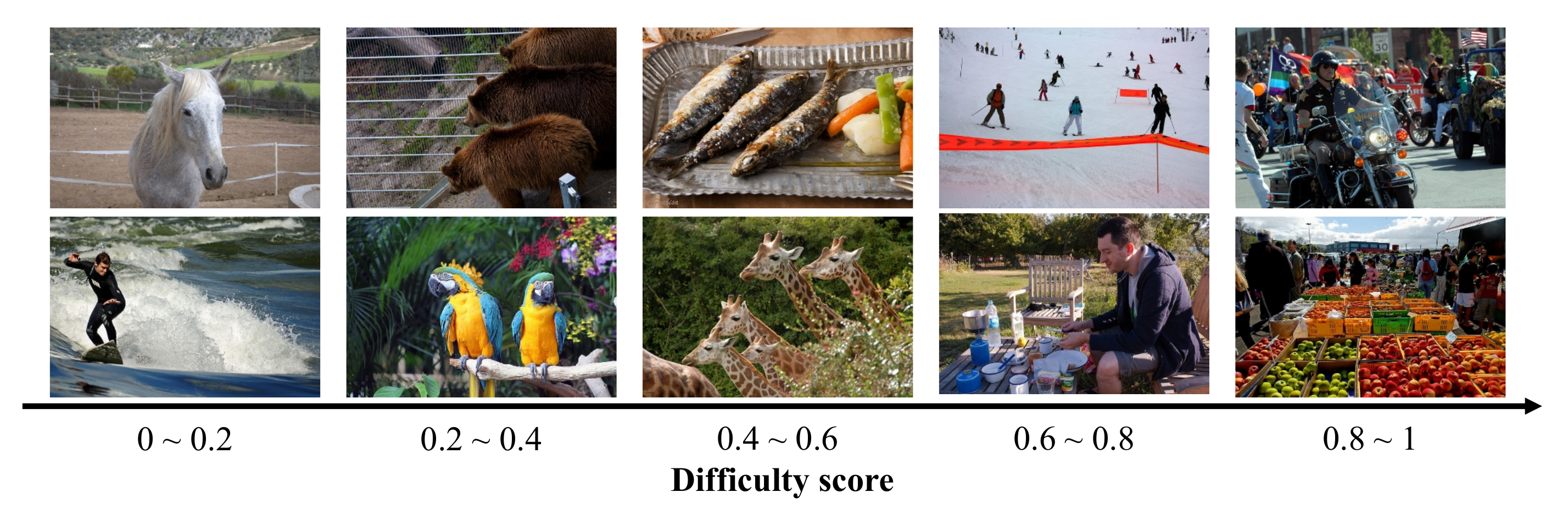}
   \vspace{-3.5mm}
   \caption{Visualization of the ``easy" and the ``hard" images. The horizontal direction corresponds to the difficulty scores predicted by our adaptive router in Dy-Mask R-CNN Swin-T.}
   \vspace{-4mm}
   \label{fig:vis}
\end{figure*}

\begin{figure}[t]
  \centering
   \includegraphics[height=0.56\linewidth]{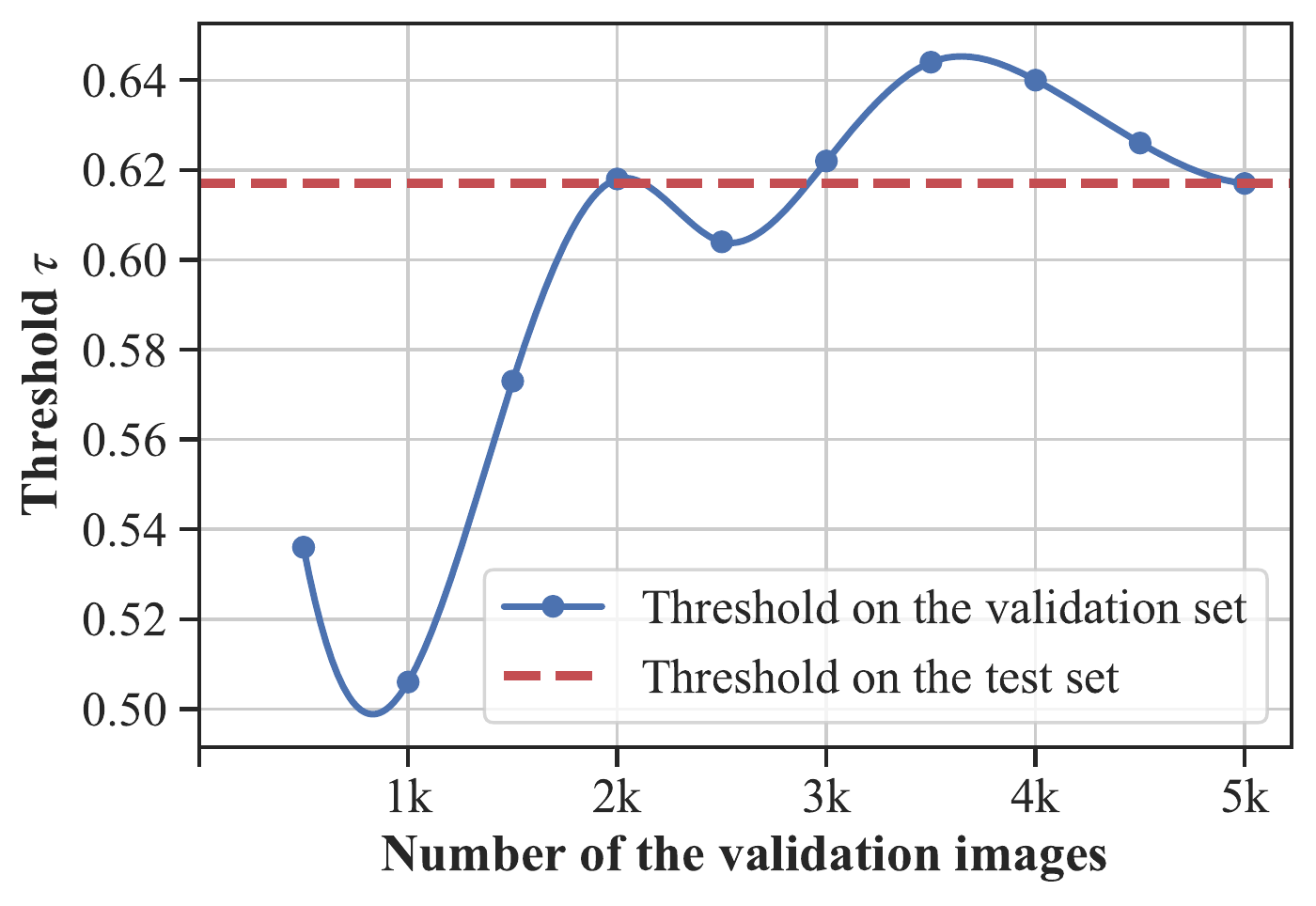}
   \vspace{-3mm}
   \caption{Comparison of the thresholds obtained from the test set and the validation sets of different sizes.}
   \vspace{-3.5mm}
   \label{fig:thres}
\end{figure}

\vspace{-4mm}
\subsubsection{Robust variable-speed inference strategy}
\vspace{-2mm}
To achieve variable-speed inference for a dynamic detector, we count the difficulty scores on the validation set and directly adopt the corresponding thresholds for the test set. This strategy requires the validation set to be large enough.
However, with custom datasets, this is not always sufficient.
To demonstrate the robustness of our variable-speed inference strategy, we analyze the impact of the validation set size on the threshold consistency between the validation set and test set. Taking the Dy-Mask R-CNN Swin-T on COCO~\cite{lin2014microsoft} dataset as an example, its threshold for 50\% quantile on the test set is 0.62. Then, we count the thresholds for 50\% quantile on the validation set of different sizes (\ie, 0.5k, 1k, $\dots$, 5k). As shown in \cref{fig:thres}, the threshold obtained from 5k validation images is consistent with the threshold of the test set, which confirms our assumption in \cref{sec:inference}. Later, as the data size decreases, the thresholds of the validation set change within a small range. However, when the data size is less than 1.5k, the threshold of the validation set and the test set will occur a large deviation~(\ie, 0.11 at 1k). Overall, our variable-speed inference strategy is stable when the validation set size is relatively sufficient~(\eg, about 2k validation images for the 20k test set on COCO~\cite{lin2014microsoft}).

\vspace{-3.5mm}
\subsubsection{Comparison with the trade-offs obtained by adjusting the input resolution}

\begin{figure}[t]
  \centering
  \includegraphics[height=0.56\linewidth]{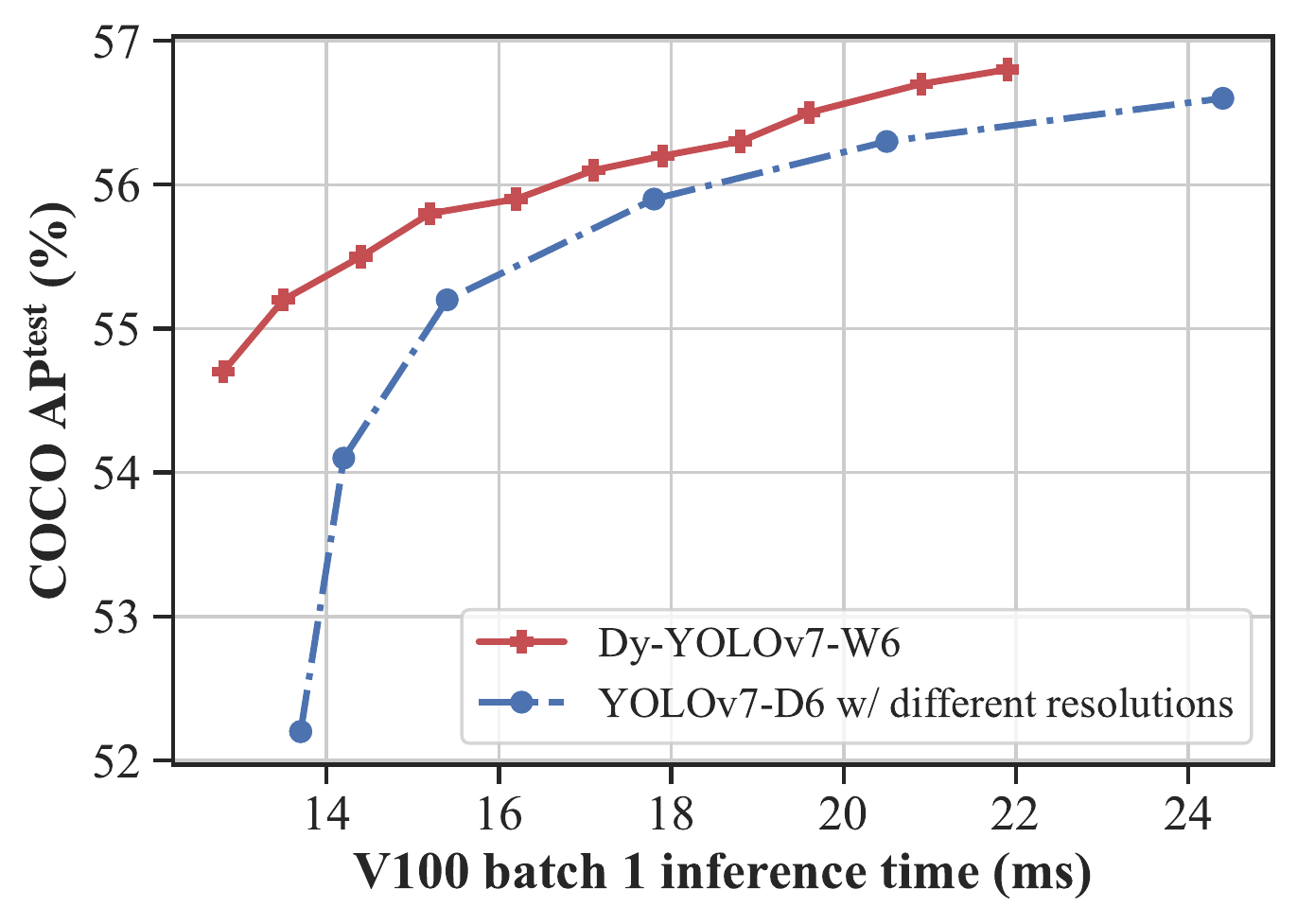}
  \vspace{-3mm}
  \caption{Comparison of the trade-offs obtained from our Dy-YOLOv7-W6 and YOLOv7-D6 with different resolutions.}
  \vspace{-3.5mm}
  \label{fig:res_trade_off}
\end{figure}

For a well-trained detector, changing its input resolution can also quickly obtain a series of accuracy-speed trade-offs. Here we compare this method with our dynamic detector. As shown in \cref{fig:res_trade_off}, we compare our Dy-YOLOv7-W6 and the YOLOv7-D6 with different input resolutions~(\ie, 640$\sim$1280), and we observe that our dynamic detector achieves better accuracy-speed trade-offs. For example, our Dy-YOLOv7-W6 achieves 55.2\% AP at 74 FPS (13.5 ms), while YOLOv7-D6 with 640 input resolution only achieves 52.2\% AP at an even slower inference speed.

\subsection{Visualization of images with different difficulty scores}
\label{sec:visualization}
We depict the images with different predicted difficulty scores in \cref{fig:vis}, ascending from left to right.
That is, the images on the left are considered as the ``easy" images, while those on the right are considered as the ``hard" images. We can observe that the ``easy" images usually contain fewer objects, with the usual camera viewpoint and the clean background. In contrast, the ``hard" images usually have more complex scenes with severe occlusion and much more small objects.

\section{Conclusion}
In this paper, we present a unified dynamic architecture for object detection, DynamicDet. We first design a dynamic architecture to support dynamic inference on mainstream detectors. Then, we propose an adaptive router to predict the difficulty score of each image and determine the inference route. With the above architecture and router, we then propose a hyperparameter-free optimization strategy with an adaptive offset to training our dynamic detectors. Last, we present a variable-speed inference strategy. With the settable threshold for dynamic inference, we can achieve a wide range of accuracy-speed trade-offs with only one dynamic detector.
Extensive experimental results demonstrate the superiority of the proposed DynamicDet in accuracy and efficiency, and new state-of-the-art accuracy-speed trade-offs are achieved.

\section*{Acknowledgments}
This work was supported by National Natural Science Foundation of China under Grant 62176007. This work was also
a research achievement of Key Laboratory of Science, Technology and Standard in Press Industry (Key Laboratory of Intelligent Press Media Technology).

\clearpage

{\small
\bibliographystyle{ieee_fullname}
\bibliography{egbib}
}

\onecolumn
\clearpage
\appendix
\vspace{-6mm}
\section{Pseudo code for dynamic detector}

We present the pseudo code of training the adaptive router on \cref{al:train}, and the dynamic inference on \cref{al:inference}.

\vspace{-2mm}
\begin{algorithm*}[h]
\small
  \SetAlgoLined
  \KwIn{The dynamic detector constructed by the first backbone $\mathcal{B}_1$, the first neck and head $\mathcal{D}_1$, the second backbone $\mathcal{B}_2$, the second neck and head $\mathcal{D}_2$, the composite connection module $\mathcal{G}$, and the adaptive router $\mathcal{R}$. The median of the training loss difference between two detectors $\Delta$. Input images $\mathbf{x}_i\in\mathbf{X}$ and the corresponding ground truths $\overline{\mathbf{y}}_i\in\overline{\mathbf{Y}}$. Training iteration $T$.}
  \For{$i= 1,\dots, T$}{
    $F_1=\mathcal{B}_1(\mathbf{x}_i);$\ \ \tcp{Extract the first multi-scale features.}
    $\mathbf{y}_1 = \mathcal{D}_1(F_1);$ \tcp{Predict the detection results by the first detector.}
    $\phi=\mathcal{R}(F_1);$\tcp{Predict the difficulty score.}
	$H = \mathcal{G}(F_1);$ \tcp{Embed the first multi-scale features.}
	$F_2 = \mathcal{B}_2(\mathbf{x}_i, H);$ \tcp{Extract the enhanced multi-scale features based on the input image and the embedding of previous multi-scale features.}

	$\mathbf{y}_2 = \mathcal{D}_2(F_2);$ \tcp{Predict the detection results by the second detector.}
	\textcolor{red}{$\mathcal{L}= ((1-\phi)(\mathcal{L}_{det}(\mathbf{y}_1, \overline{\mathbf{y}}_i) -{\Delta /2}) + \phi(\mathcal{L}_{det}(\mathbf{y}_2,\overline{\mathbf{y}}_i) +{\Delta /2}));$}\tcp{Loss.}
    update the parameters of adaptive router based on the gradient from loss $\mathcal{L}$.
  }
  \caption{Pseudo code of training the adaptive router on DynamicDet.}
  \label{al:train}
\end{algorithm*}

\vspace{-8mm}
\begin{algorithm*}[h]
\small
  \SetAlgoLined
  \KwIn{The dynamic detector constructed by the first backbone $\mathcal{B}_1$, the first neck and head $\mathcal{D}_1$, the second backbone $\mathcal{B}_2$, the second neck and head $\mathcal{D}_2$, the composite connection module $\mathcal{G}$, and the adaptive router $\mathcal{R}$. Input image $\mathbf{x}$. Threshold $\tau$.}
  \KwOut{Predicted detection results $\mathbf{y}$}
    $F_1=\mathcal{B}_1(\mathbf{x});$\ \ \tcp{Extract the first multi-scale features.}
    $\phi=\mathcal{R}(F_1);$\tcp{Predict the difficulty score.}
	\eIf{$\phi \le \tau$}{
	    \textcolor{red}{\tcp{Easy image.}}
		$\mathbf{y} = \mathcal{D}_1(F_1);$ \tcp{Predict the detection results by the first detector.}
	}{
    	\textcolor{red}{\tcp{Hard image.}}
    	$H = \mathcal{G}(F_1);$ \tcp{Embed the first multi-scale features.}
    	$F_2 = \mathcal{B}_2(\mathbf{x}, H);$ \tcp{Extract the enhanced multi-scale features based on the input image and the embedding of previous multi-scale features.}
    	$\mathbf{y} = \mathcal{D}_2(F_2);$ \tcp{Predict the detection results by the second detector.}
	}
  \caption{Pseudo code of dynamic inference on DynamicDet.}
  \label{al:inference}
\end{algorithm*}

\vspace{-4mm}
\section{Additional results}

\subsection{More comparison on real-time object detection}
We present more precision results~(\eg, AP$_\text{50}$) to compare with other real-time object detectors in \cref{tab:results_full}.


\subsection{More results for dynamic detectors}
We present more results for our dynamic detector~(\ie, Dy-YOLOv7 with / 0, / 10, \dots, / 100) in \cref{tab:results_dy}. It is observed that our dynamic detectors can obtain a wide range of trade-offs of different precision and speed by proposed variable-speed inference strategy. For instance, using the same weight with different thresholds for inference, our Dy-YOLOv7-W6 can achieve 54.7\%$\sim$56.8\% mAP with 78$\sim$46 FPS.

\renewcommand{\thetable}{4}
\begin{table*}[ht]
\small
\centering
\addtolength{\tabcolsep}{4.5pt}
\renewcommand\arraystretch{0.847}
\begin{threeparttable}
\begin{tabular}{l|rrr|r|rrrrr}
\toprule
Model                      & Size          & FLOPs         & FPS          & AP            & AP$_\text{50}$          & AP$_\text{75}$          & AP$_\text{S}$           & AP$_\text{M}$           & AP$_\text{L}$           \\
\midrule
EAutoDet-S~\cite{wang2022eautodet}                 & 640           & 24.9G           & 120\tnote{\dag}          & 40.1          & 58.7          & 43.5          & 21.7          & 43.8          & 50.5          \\
EAutoDet-M~\cite{wang2022eautodet}                 & 640           & 60.8G           & 70\tnote{\dag}           & 45.2          & 63.5          & 49.1          & 25.7          & 49.1          & 57.3          \\
EAutoDet-L~\cite{wang2022eautodet}                 & 640           & 115.4G          & 59\tnote{\dag}           & 47.9          & 66.3          & 52.0          & 28.3          & 52.0          & 59.9          \\
EAutoDet-X~\cite{wang2022eautodet}                 & 640           & 225.3G          & 41\tnote{\dag}           & 49.2          & 67.5          & 53.6          & 30.4          & 53.4          & 61.5          \\
\midrule
EfficientDet-D0~\cite{tan2020efficientdet} & 512 & 2.5G & 98\tnote{\dag} & 34.6 & 53.0 & 37.1 & - & - & - \\
EfficientDet-D1~\cite{tan2020efficientdet} & 640 & 6.1G & 74\tnote{\dag} & 40.5 & 59.1 & 43.7 & - & - & - \\
\midrule
YOLOX-S~\cite{ge2021yolox}                     & 640           & 26.8G  & 102\tnote{\dag}            & 40.5       &-&-&-&-&-  \\
YOLOX-M~\cite{ge2021yolox}                     & 640           & 73.8G  & 81\tnote{\dag}             & 47.2       &-&-&-&-&-   \\
YOLOX-L~\cite{ge2021yolox}                     & 640           & 155.6G & 69\tnote{\dag}             & 50.1       &-&-&-&-&-   \\
YOLOX-X~\cite{ge2021yolox}                     & 640           & 281.9G & 58\tnote{\dag}             & 51.5       &-&-&-&-&-   \\
\midrule
YOLOv5-N (r6.2)~\cite{2022githubyolov5}                  & 640           & 4.5G            & 200          & 28.1          & 46.2          & 29.4          & 12.8          & 31.3          & 35.4          \\
YOLOv5-S (r6.2)~\cite{2022githubyolov5}                   & 640           & 16.5G           & 196          & 37.7          & 57.3          & 40.5          & 19.8          & 41.7          & 47.4          \\
YOLOv5-M (r6.2)~\cite{2022githubyolov5}                   & 640           & 49.0G           & 137          & 45.4          & 64.3          & 49.2          & 26.3          & 49.9          & 56.4          \\
YOLOv5-L (r6.2)~\cite{2022githubyolov5}                   & 640           & 109.1G          & 114          & 49.0          & 67.5          & 53.1          & 29.8          & 53.4          & 61.2          \\
YOLOv5-X (r6.2)~\cite{2022githubyolov5}                  & 640           & 205.7G          & 100          & 50.9          & 69.2          & 55.1          & 31.9          & 55.2          & 63.6          \\
\midrule
YOLOv6-N~\cite{li2022yolov6}                   & 640           & 11.1G           & 216          & 36.4          & 51.9          & 39.2          & 15.5          & 39.5          & 50.6          \\
YOLOv6-T~\cite{li2022yolov6}                   & 640           & 36.7G           & 206          & 41.2          & 57.9          & 44.6          & 19.9          & 45.0          & 56.0          \\
YOLOv6-S~\cite{li2022yolov6}                   & 640           & 44.2G           & 184          & 43.9          & 60.9          & 47.5          & 22.2          & 47.9          & 58.9          \\
YOLOv6-M~\cite{li2022yolov6}                   & 640           & 82.2G           & 109          & 49.8          & 67.0          & 54.3          & 28.5          & 54.6          & 65.4          \\
YOLOv6-L~\cite{li2022yolov6}                   & 640           & 144.0G          & 76           & 52.3          & 69.9          & 56.8          & 31.6          & 57.2          & 67.8          \\
\midrule
PP-YOLOE+-S~\cite{xu2022pp}                & 640           & 17.4G           & 208\tnote{\dag}          & 43.9          & -             & -             & -             & -             & -             \\
PP-YOLOE+-M~\cite{xu2022pp}                & 640           & 49.9G           & 123\tnote{\dag}          & 50.0          & -             & -             & -             & -             & -             \\
PP-YOLOE+-L~\cite{xu2022pp}                & 640           & 110.1G          & 78\tnote{\dag}           & 53.3          & -             & -             & -             & -             & -             \\
PP-YOLOE+-X~\cite{xu2022pp}                & 640           & 206.6G          & 45\tnote{\dag}           & 54.9          & -             & -             & -             & -             & -             \\
\midrule
YOLOv7~\cite{wang2022yolov7}                     & 640           & 104.7G          & 114          & 51.4          & 69.7          & 55.9          & 31.8          & 55.5          & 65.0          \\
\textbf{Dy-YOLOv7 / 10}    & \textbf{640}  & \textbf{112.4G} & \textbf{110} & \textbf{52.1} & \textbf{70.5} & \textbf{56.8} & \textbf{33.3} & \textbf{55.9} & \textbf{64.7} \\
\textbf{Dy-YOLOv7 / 50}    & \textbf{640}  & \textbf{143.2G} & \textbf{96}  & \textbf{53.3} & \textbf{71.7} & \textbf{58.1} & \textbf{34.9} & \textbf{57.0} & \textbf{65.4} \\
\textbf{Dy-YOLOv7 / 90}    & \textbf{640}  & \textbf{174.0G} & \textbf{85}  & \textbf{53.8} & \textbf{72.2} & \textbf{58.7} & \textbf{35.3} & \textbf{57.5} & \textbf{66.3} \\
\textbf{Dy-YOLOv7 / 100}    & \textbf{640}  & \textbf{181.7G} & \textbf{83}  & \textbf{53.9} & \textbf{72.2} & \textbf{58.7} & \textbf{35.3} & \textbf{57.6} & \textbf{66.4} \\
\midrule
YOLOv7-X~\cite{wang2022yolov7}                   & 640           & 189.9G          & 105          & 53.1          & 71.2          & 57.8          & 33.8          & 57.1          & 67.4          \\
\textbf{Dy-YOLOv7-X / 10}  & \textbf{640}  & \textbf{201.7G} & \textbf{98}  & \textbf{53.3} & \textbf{71.6} & \textbf{58.0} & \textbf{34.2} & \textbf{57.1} & \textbf{67.1} \\
\textbf{Dy-YOLOv7-X / 50}  & \textbf{640}  & \textbf{248.9G} & \textbf{78}  & \textbf{54.4} & \textbf{72.7} & \textbf{59.3} & \textbf{36.0} & \textbf{58.0} & \textbf{67.7} \\
\textbf{Dy-YOLOv7-X / 90}  & \textbf{640}  & \textbf{296.1G} & \textbf{65}  & \textbf{55.0} & \textbf{73.2} & \textbf{59.9} & \textbf{36.6} & \textbf{58.6} & \textbf{68.2} \\
\textbf{Dy-YOLOv7-X / 100}  & \textbf{640}  & \textbf{307.9G} & \textbf{64}  & \textbf{55.0} & \textbf{73.2} & \textbf{60.0} & \textbf{36.6} & \textbf{58.7} & \textbf{68.5} \\
\midrule
\midrule
EfficientDet-D2~\cite{tan2020efficientdet} & 768 & 11.0G & 56\tnote{\dag} & 43.9 & 62.7 & 47.6 & - & - & - \\
EfficientDet-D3~\cite{tan2020efficientdet} & 896 & 25.0G & 34\tnote{\dag} & 47.2 & 65.9 & 51.2 & - & - & - \\
EfficientDet-D4~\cite{tan2020efficientdet} & 1024 & 55.0G & 23\tnote{\dag} & 49.7 & 68.4 & 53.9 & - & - & - \\
EfficientDet-D5~\cite{tan2020efficientdet} & 1280 & 135.0G & 14\tnote{\dag} & 51.5 & 70.5 & 56.1 & - & - & - \\
EfficientDet-D6~\cite{tan2020efficientdet} & 1280 & 226.0G & 11\tnote{\dag} & 52.6 & 71.5 & 57.2 & - & - & - \\
EfficientDet-D7~\cite{tan2020efficientdet} & 1536 & 325.0G & 8\tnote{\dag} & 53.7 & 72.4 & 58.4 & - & - & - \\
EfficientDet-D7X~\cite{tan2020efficientdet} & 1536 & 410.0G & 7\tnote{\dag} & 55.1 & 74.3 & 59.9 & - & - & - \\
\midrule
YOLOv5-N6 (r6.2)~\cite{2022githubyolov5}                  & 1280          & 18.4G           & 161          & 36.2          & 55.0          & 39.0          & 19.4          & 39.3          & 45.2          \\
YOLOv5-S6 (r6.2)~\cite{2022githubyolov5}                  & 1280          & 67.2G           & 152          & 44.6          & 63.9          & 48.6          & 26.4          & 48.3          & 55.1          \\
YOLOv5-M6 (r6.2)~\cite{2022githubyolov5}                  & 1280          & 200.0G          & 96           & 51.4          & 69.7          & 56.0          & 33.3          & 55.2          & 62.5          \\
YOLOv5-L6 (r6.2)~\cite{2022githubyolov5}                  & 1280          & 445.6G          & 65           & 53.8          & 71.8          & 58.5          & 36.3          & 57.6          & 65.0          \\
YOLOv5-X6 (r6.2)~\cite{2022githubyolov5}                  & 1280          & 839.2G          & 39           & 55.0          & 72.8          & 59.8          & 37.3          & 58.5          & 66.8          \\
\midrule
YOLOv7-W6~\cite{wang2022yolov7}                  & 1280          & 360.0G          & 78           & 54.9          & 72.6          & 60.1          & 37.3          & 58.7          & 67.1          \\
YOLOv7-E6~\cite{wang2022yolov7}                  & 1280          & 515.2G          & 52           & 56.0          & 73.5          & 61.2          & 38.0          & 59.9          & 68.4          \\
YOLOv7-D6~\cite{wang2022yolov7}                  & 1280          & 806.8G          & 41           & 56.6          & 74.0          & 61.8          & 38.8          & 60.1          & 69.5          \\
YOLOv7-E6E~\cite{wang2022yolov7}                 & 1280          & 843.2G          & 33           & 56.8          & 74.4          & 62.1          & 39.3          & 60.5          & 69.0          \\
\midrule
\textbf{Dy-YOLOv7-W6 / 10} & \textbf{1280} & \textbf{384.2G} & \textbf{74}  & \textbf{55.2} & \textbf{73.0} & \textbf{60.4} & \textbf{37.9} & \textbf{58.4} & \textbf{66.6} \\
\textbf{Dy-YOLOv7-W6 / 50} & \textbf{1280} & \textbf{480.8G} & \textbf{58}  & \textbf{56.1} & \textbf{73.8} & \textbf{61.4} & \textbf{39.3} & \textbf{59.3} & \textbf{66.9} \\
\textbf{Dy-YOLOv7-W6 / 90} & \textbf{1280} & \textbf{577.4G} & \textbf{48}  & \textbf{56.7} & \textbf{74.3} & \textbf{62.1} & \textbf{39.5} & \textbf{59.9} & \textbf{67.8} \\
\textbf{Dy-YOLOv7-W6 / 100} & \textbf{1280} & \textbf{601.6G} & \textbf{46} & \textbf{56.8} & \textbf{74.4} & \textbf{62.1} & \textbf{39.6} & \textbf{59.9} & \textbf{68.3} \\
\bottomrule
\end{tabular}
\begin{tablenotes}
	\footnotesize
	\item[1] The FPS marked with $\dag$ are from the corresponding papers, and others are measured on the same machine with 1 NVIDIA V100 GPU.
\end{tablenotes}
\end{threeparttable}
\caption{Comparison of the state-of-the-art real-time object detectors on COCO \textit{test-dev}.}
\label{tab:results_full}
\end{table*}

\renewcommand{\thetable}{5}
\begin{table*}[t]
\small
\centering
\addtolength{\tabcolsep}{4.5pt}
\begin{tabular}{l|rrr|r|rrrrr}
\toprule
Model                      & Size          & FLOPs         & FPS          & AP            & AP$_\text{50}$          & AP$_\text{75}$          & AP$_\text{S}$           & AP$_\text{M}$           & AP$_\text{L}$           \\
\midrule
Dy-YOLOv7 / 0      & 640  & 104.7G  & 114 & 51.1 & 69.5 & 55.6 & 31.5 & 55.2 & 64.5 \\
Dy-YOLOv7 / 10     & 640  & 112.4G  & 110 & 52.1 & 70.5 & 56.8 & 33.3 & 55.9 & 64.7 \\
Dy-YOLOv7 / 20     & 640  & 120.1G  & 106 & 52.5 & 71.0 & 57.3 & 34.1 & 56.2 & 64.9 \\
Dy-YOLOv7 / 30     & 640  & 127.8G  & 102 & 52.9 & 71.3 & 57.6 & 34.5 & 56.5 & 65.0 \\
Dy-YOLOv7 / 40     & 640  & 135.5G  & 99  & 53.1 & 71.6 & 57.9 & 34.7 & 56.8 & 65.2 \\
Dy-YOLOv7 / 50     & 640  & 143.2G  & 96  & 53.3 & 71.7 & 58.1 & 34.9 & 57.0 & 65.4 \\
Dy-YOLOv7 / 60     & 640  & 150.9G  & 93  & 53.5 & 71.9 & 58.3 & 35.1 & 57.2 & 65.5 \\
Dy-YOLOv7 / 70     & 640  & 158.6G  & 91  & 53.6 & 72.0 & 58.5 & 35.2 & 57.4 & 65.7 \\
Dy-YOLOv7 / 80     & 640  & 166.3G  & 88  & 53.7 & 72.1 & 58.6 & 35.3 & 57.5 & 66.0 \\
Dy-YOLOv7 / 90     & 640  & 174.0G  & 85  & 53.8 & 72.2 & 58.7 & 35.3 & 57.5 & 66.3 \\
Dy-YOLOv7 / 100    & 640  & 181.7G  & 83  & 53.9 & 72.2 & 58.7 & 35.3 & 57.6 & 66.4 \\
\midrule
Dy-YOLOv7-X / 0    & 640  & 189.9G  & 105 & 52.6 & 70.7 & 57.2 & 32.9 & 56.6 & 67.1 \\
Dy-YOLOv7-X / 10   & 640  & 201.7G  & 98  & 53.3 & 71.6 & 58.0 & 34.2 & 57.1 & 67.1 \\
Dy-YOLOv7-X / 20   & 640  & 213.5G  & 93  & 53.7 & 71.9 & 58.5 & 34.8 & 57.3 & 67.4 \\
Dy-YOLOv7-X / 30   & 640  & 225.3G  & 86  & 53.9 & 72.2 & 58.8 & 35.3 & 57.5 & 67.4 \\
Dy-YOLOv7-X / 40   & 640  & 237.1G  & 82  & 54.1 & 72.5 & 59.0 & 35.6 & 57.8 & 67.4 \\
Dy-YOLOv7-X / 50   & 640  & 248.9G  & 78  & 54.4 & 72.7 & 59.3 & 36.0 & 58.0 & 67.7 \\
Dy-YOLOv7-X / 60   & 640  & 260.7G  & 75  & 54.6 & 72.8 & 59.5 & 36.3 & 58.2 & 67.8 \\
Dy-YOLOv7-X / 70   & 640  & 272.5G  & 70  & 54.7 & 72.9 & 59.6 & 36.4 & 58.3 & 67.8 \\
Dy-YOLOv7-X / 80   & 640  & 284.3G  & 68  & 54.8 & 73.0 & 59.8 & 36.6 & 58.4 & 68.0 \\
Dy-YOLOv7-X / 90   & 640  & 296.1G  & 65  & 55.0 & 73.2 & 59.9 & 36.6 & 58.6 & 68.2 \\
Dy-YOLOv7-X / 100  & 640  & 307.9G  & 64  & 55.0 & 73.2 & 60.0 & 36.6 & 58.7 & 68.5 \\
\midrule
Dy-YOLOv7-W6 / 0   & 1280 & 360.0G  & 78  & 54.7 & 72.4 & 59.8 & 36.6 & 58.1 & 66.5 \\
Dy-YOLOv7-W6 / 10  & 1280 & 384.2G  & 74  & 55.2 & 73.0 & 60.4 & 37.9 & 58.4 & 66.6 \\
Dy-YOLOv7-W6 / 20  & 1280 & 408.3G  & 69  & 55.5 & 73.3 & 60.8 & 38.5 & 58.7 & 66.7 \\
Dy-YOLOv7-W6 / 30  & 1280 & 432.5G  & 66  & 55.8 & 73.5 & 61.1 & 38.8 & 58.9 & 66.7 \\
Dy-YOLOv7-W6 / 40  & 1280 & 456.6G  & 62  & 55.9 & 73.7 & 61.2 & 39.1 & 59.1 & 66.8 \\
Dy-YOLOv7-W6 / 50  & 1280 & 480.8G  & 58  & 56.1 & 73.8 & 61.4 & 39.3 & 59.3 & 66.9 \\
Dy-YOLOv7-W6 / 60  & 1280 & 505.0G  & 56  & 56.2 & 73.9 & 61.6 & 39.4 & 59.4 & 67.0 \\
Dy-YOLOv7-W6 / 70  & 1280 & 529.1G  & 53  & 56.3 & 74.0 & 61.7 & 39.4 & 59.5 & 67.1 \\
Dy-YOLOv7-W6 / 80  & 1280 & 553.3G  & 51  & 56.5 & 74.2 & 61.9 & 39.4 & 59.7 & 67.5 \\
Dy-YOLOv7-W6 / 90  & 1280 & 577.4G  & 48  & 56.7 & 74.3 & 62.1 & 39.5 & 59.9 & 67.8 \\
Dy-YOLOv7-W6 / 100 & 1280 & 601.6G  & 46  & 56.8 & 74.4 & 62.1 & 39.6 & 59.9 & 68.3 \\
\bottomrule
\end{tabular}
\caption{Detailed results of dynamic YOLOv7 models on COCO \textit{test-dev}.}
\label{tab:results_dy}
\end{table*}

\section{Additional analyses}
\subsection{Consistency of the inference time}
To further demonstrate the consistency of the inference time between the validation set and the test set, we compare the inference time of Dy-YOLOv7-W6 on these two sets. Specifically, we calculate the thresholds on the validation set of different sizes (\ie, 0.5k, 2k, 5k) and measure their inference time on the validation set and the test set. As shown in \cref{fig:rel_test_val}, it is observed that the inference time is consistent between these two sets when calculating the thresholds by 5k validation images, and is very close to the ideal case. Moreover, when the validation set's size decreases, the inference time consistency becomes slightly worse but is still acceptable.

\renewcommand{\thefigure}{11}
\begin{figure}[t]
  \centering
  \includegraphics[width=0.5\linewidth]{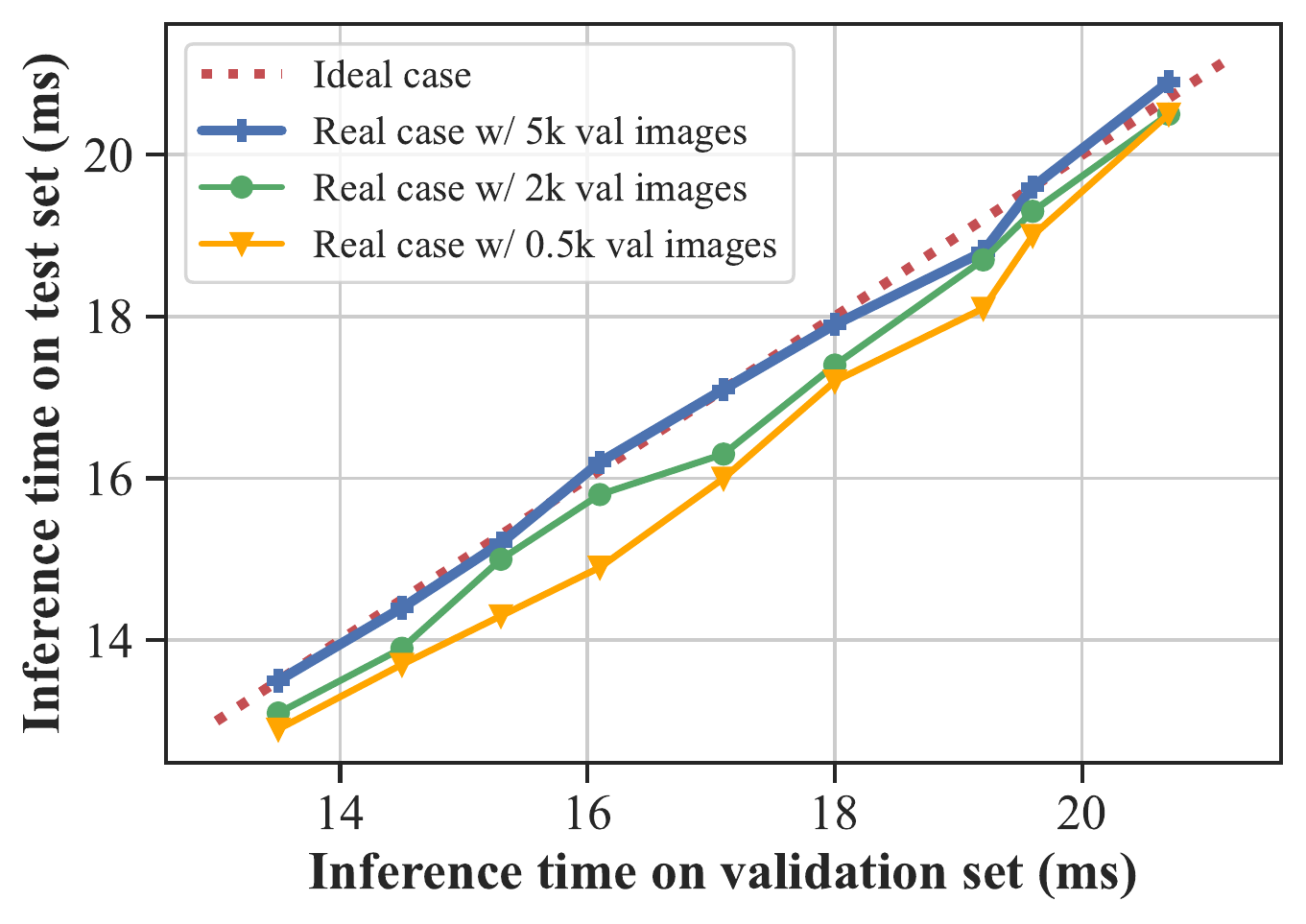}
  \vspace{-3mm}
  \caption{Comparison of the inference time on the validation set and the test set under the different thresholds obtained from the validation set with different size.}
  \vspace{-4mm}
  \label{fig:rel_test_val}
\end{figure}

\end{document}